\documentclass{egpubl}
\usepackage{egsr2026_DLOnlyTrack}
\WsPaper

\usepackage[T1]{fontenc}
\usepackage{dfadobe}

\usepackage{cite}
\BibtexOrBiblatex
\electronicVersion
\PrintedOrElectronic
\usepackage{ifpdf}
\ifpdf
  \usepackage[pdftex]{graphicx}
\else
  \usepackage[dvips]{graphicx}
\fi

\usepackage{egweblnk}
\usepackage{amsmath,amssymb,amsfonts}
\usepackage[export]{adjustbox}
\usepackage{array}
\usepackage{arydshln}
\usepackage{booktabs}
\usepackage{colortbl}
\usepackage{float,wrapfig}
\usepackage{hhline}
\usepackage{multirow}
\usepackage{tabularx}
\usepackage{newfloat}

\usepackage{bm}
\usepackage{nicefrac}
\usepackage{microtype}
\usepackage{xfrac}

\usepackage{changepage}
\usepackage{extramarks}
\usepackage{setspace}
\usepackage{soul}
\usepackage{xspace}

\usepackage{etoolbox,siunitx}
\usepackage{afterpage}
\usepackage{subcaption}

\usepackage{calc}

\usepackage{extarrows}

\usepackage{pifont}
\usepackage{enumitem}
\usepackage[ruled]{algorithm2e}

\usepackage{physics}
\usepackage{makecell}
\usepackage{cleveref}

\newcommand{\ignorethis}[1]{}

\makeatletter
\DeclareRobustCommand\onedot{\futurelet\@let@token\@onedot}
\def\@onedot{\ifx\@let@token.\else.\null\fi\xspace}

\makeatother

\newcommand*{\rom}[1]{\expandafter\romannumeral #1}

\definecolor{mydarkblue}{rgb}{0,0.08,1}
\definecolor{mydarkgreen}{rgb}{0.02,0.6,0.02}
\definecolor{mydarkred}{rgb}{0.8,0.02,0.02}
\definecolor{mydarkorange}{rgb}{0.40,0.2,0.02}
\definecolor{mypurple}{RGB}{111,0,255}
\definecolor{myred}{rgb}{1.0,0.0,0.0}
\definecolor{mygold}{rgb}{0.75,0.6,0.12}
\definecolor{myblue}{rgb}{0,0.2,0.8}
\definecolor{mydarkgray}{rgb}{0.66,0.66,0.66}

\newif\ifcolor
\colortrue

\newif\ifdraft
\draftfalse

\ifdraft
    \newcommand{\kac}[1]{{\color{magenta}\textbf{Kfir:} #1}}
    \newcommand{\ync}[1]{{\color{blue}\textbf{Yotam:} #1}}
    \newcommand{\dcc}[1]{{\color{red}\textbf{Danny:} #1}}
    \newcommand{\ygc}[1]{{\color{cyan}\textbf{Yossi:} #1}}
    \newcommand{\imc}[1]{{\color{green}\textbf{Inbar:} #1}}
    \newcommand{\qhc}[1]{{\color{teal}\textbf{Charles:} #1}}
    \newcommand{\myc}[1]{{\color{teal}\textbf{Michal:} #1}}
    \newcommand{\olc}[1]{{\color{violet}\textbf{Orly:} #1}}
    \newcommand{\ypc}[1]{{\color{red}\textbf{Yael:} #1}}

    \newcommand{\nuke}[1]{{\color{red}#1}} %
    \newcommand{\move}[1]{{\color{orange}#1}} %
    
\else
    \newcommand{\kac}[1]{}
    \newcommand{\ync}[1]{}
    \newcommand{\dcc}[1]{}
    \newcommand{\ygc}[1]{}
    \newcommand{\imc}[1]{}
    \newcommand{\qhc}[1]{}
    \newcommand{\olc}[1]{}
    \newcommand{\ypc}[1]{}
    \newcommand{\myc}[1]{}
    \newcommand{\nuke}[1]{} %
    \newcommand{\move}[1]{} %

\fi

\newif\ifcamera
\camerafalse

\ifcamera
    \newcommand{\camera}[1]{#1}
\else
    \newcommand{\camera}[1]{}
\fi

\newcommand{\vect}[1]{\boldsymbol{\mathbf{#1}}}

\newlist{todolist}{itemize}{2}
\setlist[todolist]{label=$\square$}

\usepackage{titletoc}
\usepackage[dvipsnames]{xcolor}
\usepackage{stmaryrd}

\SetAlFnt{\small}
\SetAlCapFnt{\small}
\SetAlCapNameFnt{\small}
\SetAlCapHSkip{0pt}

\newcommand{\change}[1] {\textcolor{black}{#1}}
\newcommand{\changenew}[1] {\textcolor{black}{#1}}
\newcommand{\changenewnew}[1] {\textcolor{black}{#1}}
\newcommand{\egsrcameraready}[1] {\textcolor{black}{#1}}
\renewcommand{\vect}[1] {\mathbf{#1}}

\title[ReAge3D: Re-Aging 3D Faces]%
      {ReAge3D: Re-Aging 3D Faces with View Consistency}

\author[
Libing Zeng et al.
]{
Libing Zeng$^{1}$,
Li Ma$^{2}$,
Mingming He$^{2}$,
Ning Yu$^{2}$,
Paul Debevec$^{2}$,
Nima Khademi Kalantari$^{1}$
\\
$^{1}$Texas A\&M University,
$^{2}$Netflix Eyeline Studios
}
\begin{document}

\teaser{
    \centering
    % \vspace{-0.1in}
    \includegraphics[width=0.99\linewidth]{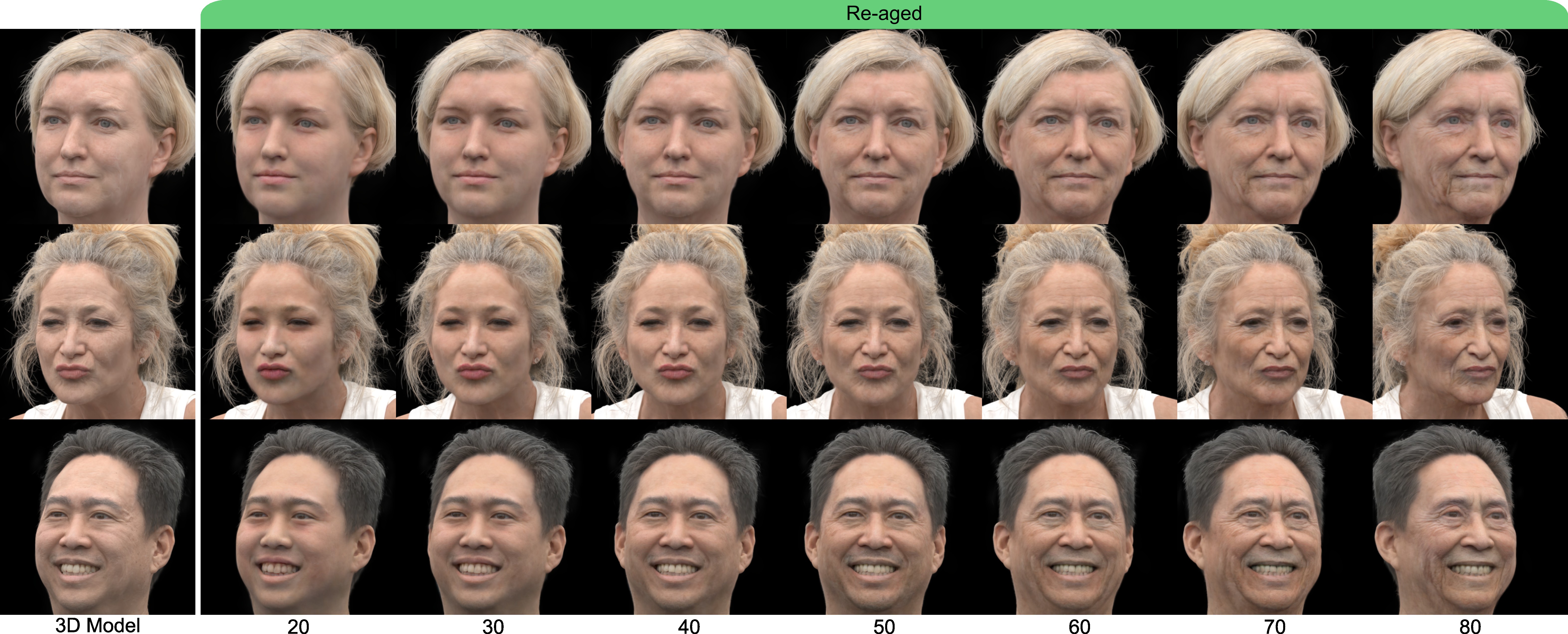}
    \vspace{-0.1in}
    \caption{Given a 3D face model, represented here using 3D Gaussian splatting, our proposed method enables precise age manipulation with fine-grained detail while maintaining multiview consistency and preserving identity. It exhibits smooth and continuous transitions across age ranges from 20 to 80 years, effectively capturing age-related changes such as the gradual formation and deepening of wrinkles. Additionally, the method is robust across a range of facial expressions (e.g. neutral, pucker, and smiles) and generalizes across different identities.
    }
    \label{fig:teaser}
    \vspace{0.1in}
}

\maketitle

\begin{abstract}
We present a novel framework for realistic and controllable 3D face re-aging which produces highly detailed, identity-preserving results. Existing 3D editing methods, while effective for coarse semantic changes, are not well suited for re-aging, as even small inconsistencies across re-aged 2D views can lead to over-smoothing of subtle but perceptually important age-related details. To address this challenge, we first introduce a 2D diffusion-based re-aging model, \textit{DiffReaging}, trained on synthetically generated image pairs. We further propose a center-out editing propagation strategy that leverages this re-aging model to reconstruct multi-view-consistent re-aged images. Specifically, starting from a re-aged frontal pivot view, we reconstruct the remaining views through warping and our proposed \textit{Masked-DiffReaging} process. By injecting existing content at every step of the diffusion process, \textit{Masked-DiffReaging} ensures that the reconstructed regions remain coherent with existing pixels. The resulting consistent set of re-aged views supervises the optimization of the re-aged 3D representation. Our method outperforms existing 3D editing techniques both visually and quantitatively, enabling smooth, fine-grained control over age transformations in 3D face models.



\end{abstract}

\vspace{-0.1in}
\section{Introduction}
\label{sec:intro}

Face re-aging -- the task of synthesizing aged or de-aged versions of an individual's face -- is a complex and compelling problem in computer graphics and vision, with diverse applications in entertainment, forensics, and biometrics. In recent years, there has been significant progress in 2D face re-aging, driven by advances in deep generative models such as generative adversarial networks (GANs)~\cite{goodfellow2014generative} and diffusion models~\cite{ho2020denoising}. These approaches typically either invert the input image into a pretrained GAN or diffusion model and perform re-aging via latent manipulation~\cite{li2021continuous, yao2021high, duong2019automatic, harkonen2020ganspace, shen2020interpreting, sam, chen2023fading, gomez2022custom}, or train a dedicated network for direct age transformation~\cite{zoss2022fran}. The latter approach, with FRAN~\cite{zoss2022fran} being the prominent example, produces high-quality re-aged images by avoiding the identity distortion commonly introduced by latent space projection. 

Despite these advances, 3D face re-aging, which aims to modify the apparent age of a 3D face model represented using techniques such as 3D Gaussian Splatting (3DGS)~\cite{3dgaussian2023} or neural radiance fields (NeRFs)~\cite{Mildenhall2020nerf}, remains relatively unexplored. A key challenge in extending 2D re-aging techniques to 3D is ensuring multi-view consistency. While not specifically designed for re-aging, several recent methods~\cite{haque2023nerf2nerf, igs2gs, chen2024gaussianeditor, chen2024dge, li2024syncnoise, wu2024gaussctrl, luo2024trame, wang2024vcgauss} perform general 3D scene editing using pretrained 2D diffusion models. Specifically, these approaches operate by editing rendered or input 2D images using the diffusion model and then use the edited images to optimize a 3D model. To encourage multi-view consistency, they typically rely on either feature-space constraints from the diffusion model~\cite{chen2024dge, wu2024gaussctrl, luo2024trame, wang2024vcgauss},  controlled noise prediction~\cite{li2024syncnoise}, or specific strategies and/or losses during optimization~\cite{haque2023nerf2nerf, igs2gs, chen2024gaussianeditor}.

Although these techniques achieve impressive results for general edits (e.g., “turn him into a clown”), they struggle to produce detailed and age-consistent re-aged models. This is primarily because age-related cues, such as wrinkles and skin texture, are subtle yet perceptually critical, and even small inconsistencies in these features across views can lead to an over-smooth 3D appearance. Furthermore, many of these approaches rely on editing models like InstructPix2Pix~\cite{brooks2023instructpix2pix}, which are not designed for fine-grained age control or identity preservation, both critical for realistic 3D face re-aging.

To overcome these challenges, we propose a novel framework that utilizes the expressiveness of diffusion models for 2D re-aging and achieves multi-view consistent re-aging directly in the image domain, as in Fig.~\ref{fig:teaser}. At the core of our method is a 2D diffusion-based re-aging model, \textit{DiffReaging}, trained to produce identity-preserving and age-consistent re-aged face images. Specifically, we propose to adapt a general-purpose editing diffusion model to the task of re-aging by fine-tuning it on synthetically re-aged images. Once trained, \textit{DiffReaging} enables high-quality re-aging of 2D images and serves as a key component in our multi-view pipeline.

Rather than re-aging each view independently, which often results in inconsistent age-related details across views, we propose to reconstruct the re-aged images by propagating the content of already re-aged images to other views to ensure multi-view consistency. Starting from a frontal pivot view, we apply \textit{DiffReaging} to generate a high-quality, identity-preserving re-aged image at the desired target age. We then warp this re-aged pivot to neighboring views using optical flow and reconstruct the missing regions by our proposed \textit{Masked-DiffReaging} process. This strategy leverages the iterative denoising nature of \textit{DiffReaging}, injecting the warped content at each timestep using a confidence mask to ensure that the reconstructed missing regions are coherent with the already existing ones. To avoid independently reconstructing overlapping content in different views, we perform the warping and \textit{Masked-DiffReaging} in a center-out manner, beginning with the pivot and progressively re-aging neighboring views in concentric layers. The resulting set of multi-view-consistent 2D re-aged images is then used to supervise the optimization of the 3D representation, with the center-out re-aging repeated periodically to refine both the supervision and the final 3D model. In summary, we make the following contributions:

\vspace{-0.05in}
\begin{itemize}
    \item We introduce a novel and practical framework for realistic 3D face re-aging, tackling a previously underexplored and challenging problem.
    \item We present a diffusion-based 2D face re-aging model capable of producing detailed re-aged images with strong identity preservation and age accuracy.
    \item We propose a content propagation strategy to produce multi-view consistent and realistic re-aging across views.
\end{itemize}
\vspace{-0.05in}

Extensive experiments show that our approach outperforms existing 3D editing methods both visually and quantitatively across a wide range of identities and expressions.
\section{Related Work}
\label{sec:relatedwork}

% \nima{It'd be great if someone could tighten this up. I will edit it later. Right now it's too long for a conference paper.}

\subsection{Face Re-Aging}
Face re-aging modifies a person’s apparent age in images or videos while preserving identity, realism, and consistency. For in-depth reviews, we refer readers to surveys on face age estimation~\cite{ramanathan2009computational,angulu2018age} and aging models~\cite{georgopoulos2018modeling,shu2016age}, and highlight key works below.
%Below, we discuss prominent lines of work on 2D face re-aging and 3D-aware methods.

\vspace{-0.05in}
\paragraph{2D Face Re-Aging.} Early methods attempt to capture age progression from limited or unpaired data, spurring the use of adversarial training~\cite{goodfellow2014generative} and cycle consistency~\cite{zhu2017unpaired}. Notable examples include partitioning data into discrete age groups~\cite{kemelmacher2014illumination} and designing conditional GANs~\cite{antipov2017face,antipov2017boosting,wang2018face}. Later methods refine these ideas by enforcing identity preservation~\cite{duong2019automatic}, disentangling identity from age~\cite{hsu2022agetransgan}, or providing continuous re-aging~\cite{li2021continuous,yao2021high}. StyleGAN-based approaches introduce latent-space editing~\cite{harkonen2020ganspace,shen2020interpreting,abdal2020styleflow}, enabling realistic transformations but often at the cost of identity or fine details~\cite{tov2021designing,roich2022pivotal,tzaban2022stitch}. More recent works incorporate text-driven diffusion~\cite{chen2023fading,li2023pluralistic} or attribute disentanglement~\cite{gomez2022custom} for refined results. A complementary approach uses a supervised encoder-decoder with synthetic training data, as in FRAN~\cite{zoss2022fran}, which encodes input faces into a 5-channel representation (including target-age masks). Attention-based strategies~\cite{Zhu2020look,Chandaliya2023AWGAN} further localize aging to specific face regions, while PADA~\cite{li2023pluralistic} integrates CLIP~\cite{radford2021learning} to manage style-content disentanglement. Despite recent progress, these 2D methods often struggle to generate high-resolution details and exhibit inconsistencies when applied to multiple views.

% \paragraph{Video Face Reaging.} 
Applying 2D re-aging frame by frame can lead to flickering or identity drift in videos. Early work~\cite{duong2019automatic} addresses sequential aging via reinforcement learning, but high-quality, temporally consistent solutions remain elusive. Some methods rely on per-frame latent-space editing in StyleGAN~\cite{yao2021latent}, but without further constraints, results exhibit temporal artifacts. Recent improvements suggest hybrid encoders~\cite{tzaban2022stitch,tov2021designing} or specialized modifications~\cite{yang2023styleganex} handle cropping and viewpoint changes, while diffusion-based editing~\cite{kim2023diffusion,preechakul2022diffusion,DBLP:journals/corr/abs-2501-03931} disentangles time-dependent features from shared identity. However, maintaining stable aging effects across diverse videos remains challenging. 

\changenewnew{More recently, several methods explore personalized or diverse face aging. TimeBooth~\cite{su2025timebooth} disentangles identity and age representations via bidirectional adversarial learning, and MyTimeMachine~\cite{qi2025mytimemachine} combines a global aging prior with a personal photo collection (${\sim}$50 images) through a StyleGAN2-based adapter network. Both methods require cross-age personal photo collections, which are unavailable in typical 3D face re-aging scenarios where the subject is captured at a single point in time. Moreover, their reliance on GAN-based architectures makes it difficult to extend them to masked inpainting, which is essential for propagating re-aged content across views. Concurrently, Cradle2Cane~\cite{liu2025cradle2cane} proposes a two-pass diffusion framework that decouples age accuracy and identity preservation, and the Aging Multiverse~\cite{gong2025agingmultiverse} generates condition-aware aging trees with diverse trajectories using training-free diffusion. While these methods advance 2D re-aging quality, they all operate without multi-view consistency guarantees. Independently editing multiple views of a 3D face leads to inconsistent age-related details across views, causing over-smoothing when optimizing the 3D representation. Our \textit{DiffReaging}, by contrast, is designed as a diffusion-based model that naturally extends to \textit{Masked-DiffReaging}, enabling pixel-level propagation of re-aged content across views with precise numerical age control.
}
%A more direct, supervised approach is explored in~\cite{muqeet2023video}, which highlights the importance of preserving both identity and temporal coherence.

\vspace{-0.05in}
\paragraph{3D-Aware Face Re-Aging.} Recent trends aim to incorporate 3D structures into 2D models for more robust view synthesis. 
%Neural radiance fields~\cite{Mildenhall2020nerf} have been adapted to face modeling~\cite{Zhang2022fdnerf,Hong2022headnerf}, and 
3D-aware GANs~\cite{chan2021,Gu2022stylenerf,Xue2022giraffehd} achieve high-quality viewpoint synthesis, but GAN-based 3D inversions~\cite{Karras2019stylegan2,Yuan2023makeencoder,Bhattarai2023triplanenet} can miss fine details. Diffusion models better preserve subtle facial features~\cite{chen2023face} but struggle with view consistency, motivating integration with 3D face structures~\cite{wahid2024diffage3d,DBLP:conf/eccv/OstrekT24}. DiffAge3D~\cite{wahid2024diffage3d} uses 3D-aware GAN priors and diffusion-based refinements for faithful multi-view re-aging but depends on a large-scale 3D-aware dataset. 
\egsrcameraready{AgeTrans3D~\cite{li2024age} performs 3D-aware re-aging by uplifting the result of a 2D re-aging diffusion model, together with the simulation of biophysical skin properties~\cite{iglesias2015biophysically}. However, its 2D re-aging model is trained from scratch as a low-resolution pixel-space diffusion model, which limits its performance.}
Another group of 3D generative models~\cite{hong2023lrm,xu2024instantmesh,SHI2023mvdream,liu2023one,liu2024one} can reconstruct 3D models from single or sparse views by leveraging multi-view consistency in diffusion models. While they achieve consistent generation across views, their fidelity and novel-view synthesis quality remain limited.

\vspace{-0.05in}
% \paragraph{Data Scarcity.} Collecting extensive, longitudinal face data with accurate age labels is challenging~\cite{ramanathan2009computational,shu2016age,angulu2018age}. Many datasets rely on age estimators or manual annotation~\cite{karras2019style,rothe2018deep,zhang2022learning,liu2021learning}, which are inaccurate and lack paired samples of the same individual at different ages. Synthetic augmentation helps address these gaps~\cite{zheng2017cross,duong2019automatic,sam} and has been proven effective for production-level face re-aging~\cite{zoss2022fran}. However, multi-view paired aging data is scarce and difficult to produce; we use 3D representations to propagate re-aging effects from frontal to other views.

%Nonetheless, robust supervised learning of continuous age progression remains a major challenge, especially for video. Approaches such as \cite{muqeet2023video} and \cite{wahid2024diffage3d} underscore the need for novel benchmarks that capture realistic temporal or multiview data.

% \NING{Please bridge the above related work to our work.}
% \mingming{Try to shorten the related work:
% \begin{enumerate}
%     \item Merge video-based reaging into 2D reaging.
%     \item The issue related to dataset can be merged into 2D or 3D reaging to save some space.
% \end{enumerate}}

\subsection{3D Representation and Editing}

% \li{Diffusion-based 3D Editing}
% \mingming{If renamed to Diffusion-based 3D Editing, do we also need to review 3D representations and other non-diffusion-based methods?}

Recent advances in 3D representations, such as neural radiance fields (NeRFs)~\cite{Mildenhall2020nerf} and 3D Gaussian Splatting (3DGS)~\cite{3dgaussian2023}, have significantly transformed the field of modeling complex 3D scenes, enabling free-viewpoint photorealistic rendering. A range of subsequent works further improve the performance of high-quality 3D reconstruction~\cite{jain2021putting,yu2021pixelnerf,barron2021mipnerf,chen2022mobilenerf,muller2022instant,chen2022tensorf,barron2023zipnerf,DBLP:conf/siggrapha/ZhangLWW022,DBLP:conf/cvpr/CharatanLTS24,DBLP:conf/cvpr/LeeRSKP24,DBLP:conf/cvpr/NiedermayrSW24,DBLP:conf/cvpr/YanLCL24,DBLP:conf/cvpr/YuCHS024,DBLP:conf/eccv/ChenXZZPGCC24} and inspire applications like 3D asset creation, often combining with diffusion models~\cite{DBLP:conf/iclr/PooleJBM23,DBLP:conf/iclr/TangRZ0Z24,DBLP:conf/cvpr/YiFWWX000W24}.

Editing 3D representations such as NeRFs and 3DGS remains challenging. The implicit nature of NeRF-based representations often necessitates global editing through text prompts~\cite{DBLP:journals/tvcg/WangJCHCL24,brooks2023instructpix2pix,DBLP:conf/nips/DongW23} or reference images~\cite{DBLP:conf/cvpr/HuangHYL022,DBLP:conf/cvpr/Bao0YFYB0C23}. 
%In contrast, the explicit structure of 3DGS offers greater efficiency of training and rendering and also enables better editability, making it a preferred choice for 3D manipulation. We also adopt this representation for human face reconstruction and manipulation. 
As an explicit representation, 3DGS supports direct edits by modifying Gaussian parameters. Prior methods~\cite{DBLP:conf/cvpr/YuJMNJ24,DBLP:conf/cvpr/HuangSYLC024} enable manipulation of 3DGS scene elements and extend this capability to motion control in dynamic scenes. However, fine-grained editing, such as re-aging, cannot be achieved by explicit editing and often benefits from the integration of text- or image-guided methods and diffusion models.

Text-guided 3DGS editing methods typically incorporate 3DGS with text embeddings from large vision-language models, such as CLIP~\cite{radford2021learning}, or utilize pretrained diffusion models~\cite{ho2020denoising,DBLP:conf/cvpr/AvrahamiLF22,brooks2023instructpix2pix,DBLP:conf/cvpr/KawarZLTCDMI23,DBLP:conf/cvpr/Chen00L24} to edit the training images based on text prompts. Some approaches~\cite{DBLP:journals/corr/abs-2403-05154,DBLP:conf/cvpr/Chen00L24} leverage Score Distillation Sampling (SDS)~\cite{DBLP:conf/iclr/PooleJBM23} to guide image editing, while others~\cite{DBLP:conf/cvpr/ZhouCJFZXCYWK24,chen2024gaussianeditor,DBLP:conf/cvpr/WangF0X024,DBLP:journals/corr/abs-2408-00083,igs2gs,DBLP:journals/tog/ZhuangKCLLS24} utilize image-conditioned diffusion models to improve spatial control. To enable localized editing, some works~\cite{DBLP:conf/cvpr/ZhouCJFZXCYWK24,chen2024gaussianeditor,DBLP:conf/cvpr/WangF0X024,DBLP:journals/corr/abs-2408-00083} propose first segmenting or tracking regions of interest and then applying text-guided modifications. % These include tasks such as changing object colors, removing or replacing objects, and modifying textures based on descriptions. 
Another line of research on global editing, such as Instruct-GS2GS~\cite{igs2gs} and TIP-Editor~\cite{DBLP:journals/tog/ZhuangKCLLS24}, iteratively edits input images using image-conditioned diffusion models like Instruct-Pix2Pix~\cite{brooks2023instructpix2pix} while optimizing the underlying 3D scene. Although these methods can produce edits faithful to text prompts, they often suffer from multi-view inconsistency and blurry reconstruction due to discrepancies across guidance images. To address view inconsistency, recent approaches integrate 3D geometry cues, such as epipolar constraints~\cite{chen2024dge}, depth supervision~\cite{li2024syncnoise,wu2024gaussctrl, lee2024editsplat}, or multi-view cross-attention maps~\cite{wang2024vcgauss, lee2024editsplat, wynn2025morpheus} into the diffusion models, or align attention-based latent codes using reference views~\cite{wu2024gaussctrl}. However, these constraints are typically applied in latent space and can not ensure pixel-level consistency. We propose an optimization framework that combines an image-conditioned re-aging model with 3DGS reconstruction to achieve fine-grained view-consistent results.

\section{Preliminaries}
\label{sec:preliminary}

% \mingming{Consider to remove this section by shortening it to one or two paragraphs and merge them with the previous section.}

% ----------------------------------------------------------------------------
\begin{figure*}[t]
\vspace{-0.1in}
\centering
\includegraphics[width=1.0\linewidth, scale=1.0, angle=-0]{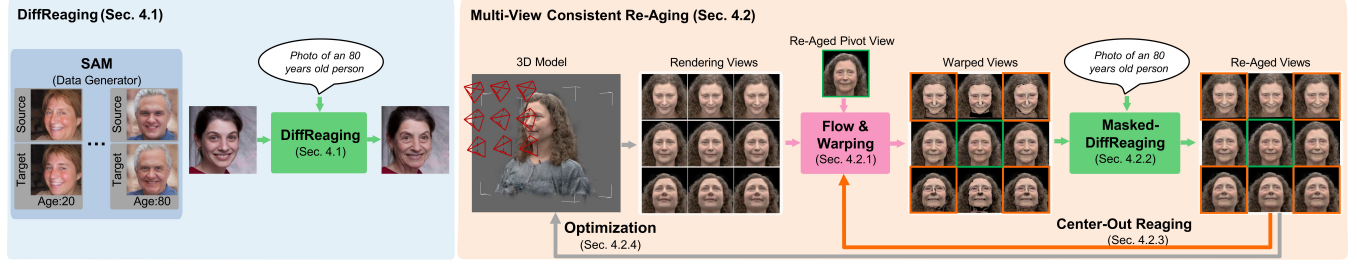}
\vspace{-0.2in}
\caption{Overview of Our 3D Face Re-Aging Framework. We first train a diffusion-based re-aging model, \textit{DiffReaging}, on a synthetic dataset. To extend its re-aging capability to 3D, we introduce a multi-view consistent editing framework. Starting with the pivot view, we re-age it using \textit{DiffReaging} and warp the results to other views. The \textit{Masked-DiffReaging} model, adapted from \textit{DiffReaging} with additional mask constraints, is then used to reconstruct occluded regions. Using a center-out re-aging strategy, we progressively re-age surrounding views. Finally, all re-aged images are used to update the 3D model.
}
% \caption{Overview of our 3D Face Reaging pipeline. Given a 3D Gaussian Splatting of a face, we iteratively optimize to its re-aged counterpart. In each iteration, we render and edit multiple views using our proposed InstructReaging model in a manner that enforces multi-view consistency. The edited views are then used to update the 3D Gaussian Splatting. \li{add section to InstructReaging and MV-Consistent Editing, change color and pattern to instruct re-aging and prompt condition, mark $t$ and $t+1$ to 3D model, add 3D model at $t=0$ and $t=T$.}}
\vspace{-0.2in}
\label{fig:overview}
\end{figure*}
% ------------

\paragraph{3D Gaussian Splatting (3DGS)} Although our technique is agonistic to the choice of 3D representation, we implement our system using 3DGS \cite{3dgaussian2023}, which explicitly represents a 3D scene as a collection of $N$ anisotropic Gaussians $\mathcal{G} = \{ \mathcal{G}_i \}_{i=1}^N$. Each Gaussian primitive in 3DGS is defined with a set of optimizable parameters, $\mathcal{G}_i = \{ \vect{p}_i, \Sigma_i, \vect{c}_i, \alpha_i \}$, consisting of 3D position $\vect{p}_i$, anisotropic covariance matrix $\Sigma_i$, RGB color $\vect{c}_i$, and opacity $\alpha_i$. Given the representation and a camera pose $v$, 3DGS uses a fast differentiable renderer $R$ to produce an image $\vect{I} = R(\mathcal{G}, \vect{v})$. We demonstrate the results of our approach on 3D faces, represented with 3DGS. 
% However, our technique is general and works with other representations such as NeRFs.

\vspace{-0.05in}
\paragraph{Diffusion Models} The diffusion model generates an image $\vect{x}_0$ by progressively denoising pure noise through a series of intermediate steps and consists of a forward and reverse process. In the forward process, noise $\epsilon$ with progressively increasing strength is added to clean data $\vect{x}_0$ to produce noisy images at different time steps $t \in \{0, \dots, T\}$, following the formulation $\vect{x}_t = \sqrt{\bar{\alpha}_t} \vect{x}_0 + \sqrt{1 - \bar{\alpha}_t} \epsilon$, where $\bar{\alpha}_t$ is the cumulative product of noise scaling factors across time steps. The reverse process aims to invert the forward process by denoising $\vect{x}_T$ back to $\vect{x}_0$ using the learned denoising function $\epsilon_\theta$, where $\theta$ denotes the model parameters. This denoising diffusion model takes as input the current image $\vect{x}_t$ and typically a text prompt $\vect{c}$ and estimates the noise, \textit{i.e.}, $\hat{\epsilon} = \epsilon_\theta(\vect{x}_t, \vect{c})$. The model is trained by optimizing the following objective:

\vspace{-0.15in}
\begin{equation}
\label{eq:loss_noise}
    \mathcal{L}_{\text{diff}} = \mathbb{E} \left[ \Vert \epsilon_{\theta}(\vect{x}_t, \vect{c}) - \epsilon \Vert^2 \right].
\end{equation}
\vspace{-0.15in}

In our method, we build our 2D re-aging model upon a variant of diffusion models, known as latent diffusion models (LDMs)~\cite{rombach2022high}. LDMs follow the same overall procedure but operate in the latent space of a variational autoencoder (VAE) to reduce computational and memory costs.

\section{Method}
\label{sec:method}

Given a 3D face representation $\mathcal{G}_o$, our objective is to produce a re-aged 3D face $\mathcal{G}_r$ that preserves the identity of the original face and accurately reflects the target age, as illustrated in Fig.~\ref{fig:teaser}. Note that our approach primarily focuses on re-aging the face, as other age-related features, such as hair,  clothing, and accessories, are typically handled through practical effects in production~\cite{zoss2022fran}.

A straightforward way to tackle this problem is to first set up a set of $N$ predefined camera poses $\vect{v}_i$ and render images using the original 3D model $\{\vect{I}_i = R(\mathcal{G}_o, \vect{v}_i)\}_{i = 1}^N$. An existing 2D re-aging approach can then be used to produce the corresponding re-aged images $\vect{L}_i$. These re-aged images can, in turn, be used to optimize a new re-aged 3D representation. This simple approach, however, leads to unsatisfactory results, as the 2D re-aging model produces images that lack 3D consistency, i.e., it adds age-related details, such as wrinkles, inconsistently across different views. Using such images to optimize the 3D model thus leads to overblurring of the details.

We address this challenge by proposing a novel framework that produces re-aged, highly detailed 3D models with the age-consistent appearance. The key idea of our approach is to ensure multi-view consistency of age-related details by propagating the content of a re-aged image to other views, rather than re-aging each view independently. To this end, we introduce \textit{DiffReaging} (Sec.~\ref{sub:diffreaging}), a 2D diffusion-based re-aging model capable of generating age-consistent, identity-preserving images. Starting from a \emph{pivot} view (frontal in our implementation), we apply \textit{DiffReaging} and propagate the re-aged content to the neighboring views through warping (Sec.~\ref{subsub:flowwarping}). We then reconstruct the missing regions in these views through our proposed \textit{Masked-DiffReaging} strategy (Sec.~\ref{subsub:Masked-DiffReaging}). Crucially, \textit{Masked-DiffReaging} leverages the iterative denoising process of diffusion models to inject known content at every step, allowing the model to reconstruct missing regions coherently with the existing ones. To avoid redundant or inconsistent reconstructions of the same 3D region across multiple views, we perform warping and \textit{Masked-DiffReaging} in a center-out manner, starting from the pivot and progressively expanding to surrounding views (Sec.\ref{subsub:center-out-reaging}). The resulting set of 2D re-aged images is then used to supervise the optimization of a 3D face model (Sec.~\ref{subsub:optimization}), with periodic updates to the supervision using the center-out re-aging strategy. An overview of our framework is shown in Fig.~\ref{fig:overview}, and the following sections detail each component.

\changenew{Note that our approach first generates view-consistent re-aged targets, which are then used to supervise a 3D face model. As a result, the framework is agnostic to the underlying 3D representation and supports optimization with any differentiable renderer (e.g., NeRF~\cite{Mildenhall2020nerf} or 3DGS~\cite{3dgaussian2023}). In this work, we primarily demonstrate results using 3DGS due to its computational efficiency.}

\vspace{-0.1in}
\subsection{DiffReaging}
\label{sub:diffreaging}

Given an input image $I$ and a target age, our goal is to generate a re-aged image $L$ of the same person at the specified age. Inspired by the existing diffusion-based image-to-image translation methods~\cite{brooks2023instructpix2pix}, we propose to model the process using a conditional diffusion model. Specifically, our model, dubbed \textit{DiffReaging}, takes the input image $I$ and the target age in the form of a text prompt, ``Photo of a \{target age\} years old person.'' as the condition and perform the iterative denoising to produce the re-aged image $\vect{x}_0$ from pure noise $\vect{x}_T$. Since we use latent diffusion models, the denoising process occurs in the latent space and the final re-aged image $\vect{L}$ is obtained by decoding the clean latent $\vect{x}_0$.

To train our model, we require triplets consisting of pixel-aligned source (input) images, corresponding re-aged (target) images, and the associated target age. We construct this dataset using SAM~\cite{sam}, a GAN-based age-conditioned face image generator. Although SAM often fails to preserve the identity of the input image in its re-aged outputs, prior work~\cite{zoss2022fran, li2024age} has shown that the identities across these re-aged outputs are internally consistent. Based on this observation, we use the re-aged images generated by SAM to form our training triplets. Specifically, we select 2,000 identities from the FFHQ dataset~\cite{karras2019style} and use SAM to re-age each image to eight target ages ranging from 10 to 80 years. 

During training, we randomly sample two of the eight re-aged images for each identity. One image is used as the source $\vect{I}$, while the other $\vect{L}^\prime$ serves as the basis for the target. To ensure that the model focuses on learning age-related changes primarily in the facial region, we compute a face mask $\vect{M}$ using BiSeNetV2~\cite{yu2021bisenet} and form the final target as $\vect{L} = (1 - \vect{M}) \odot \vect{I} + \vect{M} \odot \vect{L}^\prime$ (see Fig.~\ref{fig:overview} left). This blending retains the background and non-face regions of the source while applying the age transformation only to the face. Note that the original images from the FFHQ are only used to produce the re-aged images and are not part of our training data.

We base our model on pre-trained InstructPix2Pix~\cite{brooks2023instructpix2pix} and fine-tune it on our re-aging dataset using standard diffusion loss in Eq.~\ref{eq:loss_noise}. Note that the diffusion model in this case takes the input image $\vect{I}$ in addition to the text prompt $\vect{c}$ as the condition, i.e., $\epsilon_\theta(\vect{x}_t, \vect{c}, \vect{I})$. Once trained, \textit{DiffReaging} is able to produce detailed, age-consistent, and identity-preserving re-aged results that are better than state-of-the-art 2D re-aging methods (Fig.~\ref{fig:reaging_diffusion_2d2} and Table~\ref{tab:age_id_2d}).

% ----------------------------------------------------------------------------
\begin{figure}[t]
% \vspace{-0.1in}
\centering
\includegraphics[width=1.0\linewidth, scale=1.0, angle=-0]{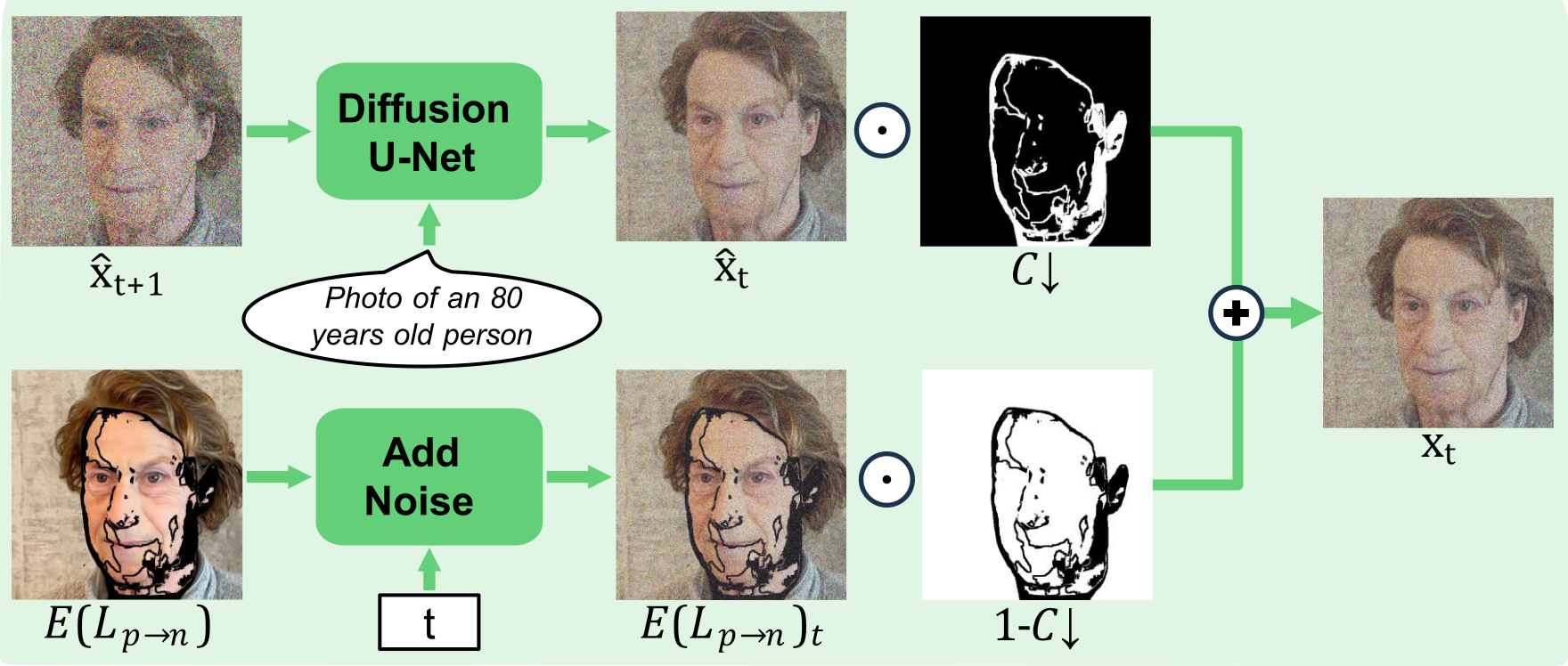}
\vspace{-0.2in}
\caption{Masked-DiffReaging. 
% Given a partially re-aged image that aligns with a target age, the objective of the Masked-DiffReaging is to fill the missing regions (represented as black holes) with pixels that are consistent with both source image and the target age.
Given a warped re-aged image with missing regions (represented as black holes), \textit{Masked-DiffReaging} inject the incomplete result into the diffusion process of \textit{DiffReaging} at every step $t$ with a confidence mask. While all operations are conducted in latent space, we show the original image here for illustration purposes.
}
\vspace{-0.2in}
\label{fig:inpainting}
\end{figure}

\subsection{Multi-View Consistent Re-Aging}
\label{sub:multiview_reaging}

Our goal is to use \textit{DiffReaging} to produce a detailed re-aged 3D model $\mathcal{G}_r$ from the original representation $\mathcal{G}_o$. To this end, we render a set of $N$ images $\vect{I}_i = R(\mathcal{G}_o, \vect{v}_i)$ by placing virtual cameras in a grid pattern over a viewing hemisphere centered on the face. Rather than generating the re-aged images $\vect{L}_i$ independently, our key idea is to enforce multi-view consistency by first re-aging the frontal \emph{pivot} image $\vect{I}_p$ and then reconstructing the remaining views by propagating the re-aged content outward from the pivot in a center-out fashion. These re-aged images are then used to optimize a detailed and age-consistent 3D model.

We begin by describing our warping module and proposed \textit{Masked-DiffReaging} technique, which both are key to propagating details across views. We then explain how these components are used together to reconstruct the re-aged views in a center-out manner. Finally, we detail our optimization strategy.

\subsubsection{Flow and Warping}
\label{subsub:flowwarping}

Given the pivot image $\vect{I}_p$, rendered using the frontal camera pose $\vect{v}_p$ and the original 3D model $\mathcal{G}_o$, we first apply \textit{DiffReaging} to produce a re-aged pivot image $\vect{L}_p$. Our goal is then to warp the content of the re-aged pivot to the neighboring views. To do so, we compute optical flow from each neighboring view $n$ to the pivot view, denoted as $\vect{F}_{n \shortrightarrow p}$, and then align the pivot image to the neighboring view through backward warping:

\vspace{-0.15in}
\begin{equation}
\label{eq:warp}
   \vect{L}_{p \shortrightarrow n} = \mathcal{W}(\vect{L}_{p}, \vect{F}_{n \shortrightarrow p}),
\end{equation}
\vspace{-0.15in}

\noindent where $\mathcal{W}$ denotes a backward warping function implemented via bilinear interpolation.

The main challenge here is that since we aim to warp the re-aged pivot, the optical flow must be computed between the re-aged neighboring and pivot views. However, the re-aged neighboring view is precisely what we are trying to reconstruct and is therefore unavailable. To overcome this, we propose computing the optical flow between the rendered neighboring and pivot images using the re-aged 3D model $\mathcal{G}_r$:

\vspace{-0.15in}
\begin{equation}
\label{eq:flow}
   \vect{F}_{n \shortrightarrow p} = \mathcal{O}\left(R(\mathcal{G}_r, \vect{v}_n), R(\mathcal{G}_r, \vect{v}_p)\right),
\end{equation}
\vspace{-0.15in}

\noindent where $\mathcal{O}$ is the optical flow operator, and we use SEA-RAFT~\cite{wang2024searaft} in our implementation. We initialize $\mathcal{G}_r$ with $\mathcal{G}_o$ at the start of optimization (Sec.~\ref{subsub:optimization}). As optimization progresses, $\mathcal{G}_r$ moves towards a re-aged representation, leading to improved optical flow, which in turn enhances the 3D model (see Fig.~\ref{fig:flow_warping_improvement}).

While optical flow provides dense correspondences, not all flow vectors are reliable due to occlusions and estimation errors. To address this, we compute a binary confidence mask $C$ using the forward-backward consistency check, retaining only the warped pixels associated with reliable flow estimates. An example of the warped image $\vect{L}_{p \shortrightarrow n}$ and its corresponding confidence mask $C$ is shown in Fig.~\ref{fig:inpainting}. In the next section, we describe our strategy for reconstructing the missing regions in $\vect{L}_{p \shortrightarrow n}$ to obtain the complete re-aged neighboring view $\vect{L}_n$.

\subsubsection{Masked-DiffReaging}
\label{subsub:Masked-DiffReaging}

Our goal is to reconstruct the missing regions with details that are consistent with the existing content in the warped image. While directly applying \textit{DiffReaging} to the neighboring image $\vect{I}_n$ yields a high-quality re-aged result, the age-related details may not align with those in the warped image $\vect{L}_{p \shortrightarrow n}$.

Inspired by Lugmayr et al.~\cite{lugmayr2022repaint}, we address this issue by leveraging the iterative denoising nature of the \textit{DiffReaging}'s diffusion process. As illustrated in Fig.~\ref{fig:inpainting}, in our method, dubbed \textit{Masked-DiffReaging}, we inject the warped image into the diffusion output at each time step $t$ using the confidence mask as follows:

\vspace{-0.15in}
\begin{equation}
\label{eq:inpaiting_unet}
\vect{x}_{t} = C\hspace{-0.03in}\downarrow \odot \ \hat{\vect{x}}_{t} + (1 - C\hspace{-0.03in}\downarrow) \odot E(\vect{L}_{p \shortrightarrow n})_{t}
\end{equation}
\vspace{-0.15in}

\noindent where $C\hspace{-0.03in}\downarrow$ is the confidence mask downsampled to the latent resolution, $\hat{\vect{x}}_{t}$ is the denoised latent at step $t$, and $E$ is the VAE encoder that maps an image to its latent representation. Moreover, $E(\vect{L}{p \shortrightarrow n})_t$ denotes the latent of the warped image after adding noise at timestep $t$. Finally, $\vect{x}_t$ is the blended latent at step $t$, used as the input to the diffusion model at the next time step. 

Starting from pure noise $\vect{x}_T$, we perform the iterative denoising and blending until we obtain the clean re-aged latent $\vect{x}_0$. This latent is then passed through the VAE decoder to generate the final re-aged image at the neighboring view, $\vect{L}_n$.

\begin{figure}[t]
\vspace{-0.05in}
\centering
\includegraphics[width=0.65\linewidth, scale=1.0, angle=-0]{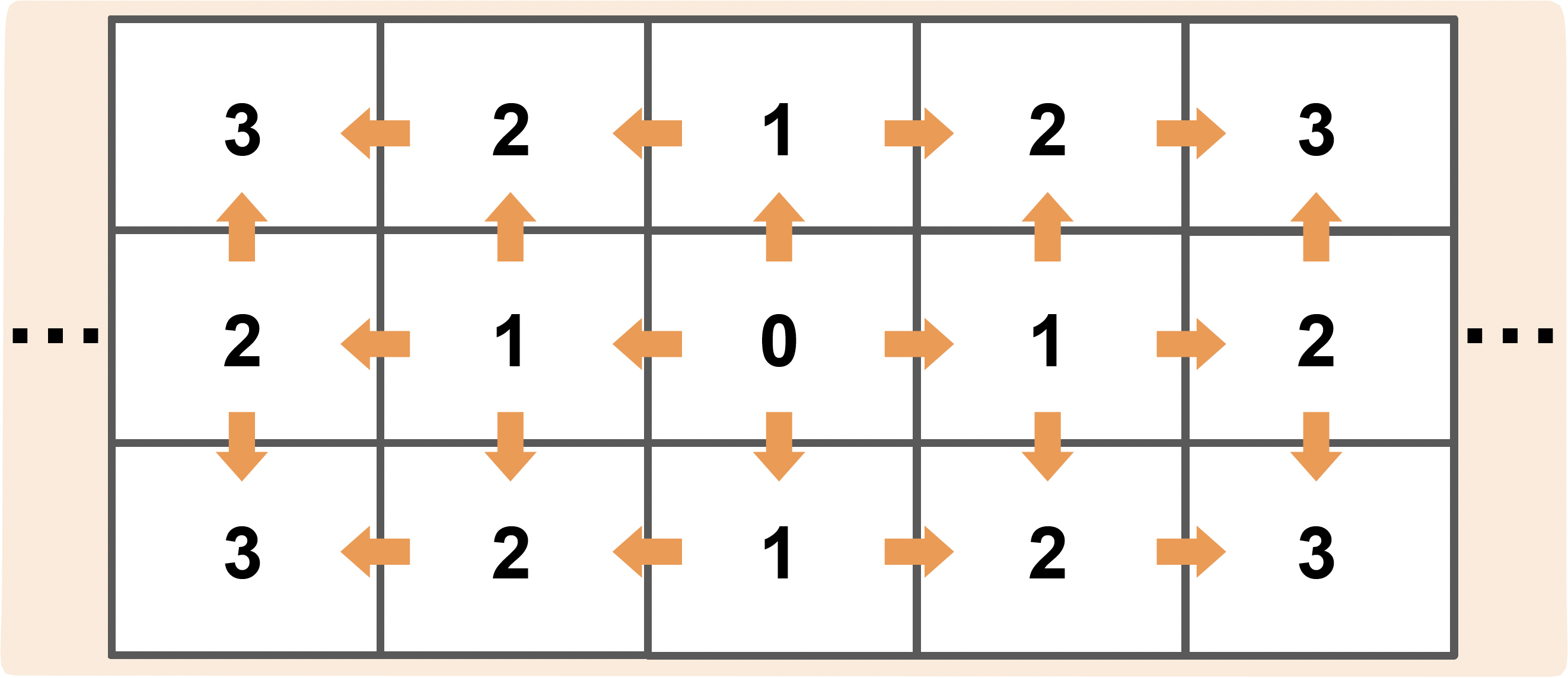}
\vspace{-0.1in}
\caption{Center-Out Re-Aging. Progressive reconstruction of multi-view images starting from a central pivot view (labeled as 0). Neighboring views are reconstructed outward in concentric layers. At each iteration, we apply warping and \textit{Masked-DiffReaging} to generate the next set of views. The labels indicate the corresponding iteration steps.}
\vspace{-0.15in}
\label{fig:flow_pattern}
\end{figure}

% % ----------------------------------------------------------------------------

\subsubsection{Center-Out Re-Aging}
\label{subsub:center-out-reaging}

While the proposed warping and \textit{Masked-DiffReaging} can be used to reconstruct all non-pivot views directly from the pivot image, doing so results in independently filling the unmasked areas in each non-pivot view. This, in turn, leads to inconsistencies in the generated age-related details across views that correspond to the same 3D region and negatively impacts the quality of the results, as shown in Fig.~\ref{fig:ablation} (``+Pivot+W''). To address this, we propose a center-out re-aging strategy, where we begin with the pivot and iteratively propagate the content outward by using already reconstructed neighbors to guide the reconstruction of adjacent views. In this strategy, illustrated in Fig.~\ref{fig:flow_pattern}, we start from the pivot view at the center (marked as 0) and progressively reconstruct neighboring views in concentric layers. At each iteration, we perform warping and \textit{Masked-DiffReaging} to reconstruct the next set of views, moving outward from the pivot.

Specifically, in the first iteration, we reconstruct the four immediate neighbors (labeled 1) by warping the pivot view and applying \textit{Masked-DiffReaging}. In the second iteration, we reconstruct the next ring of neighbors (labeled 2), and so on. For each target view, we combine information from multiple previously reconstructed views. As indicated by the arrows in Fig.~\ref{fig:flow_pattern}, we warp the pivot view (not shown to reduce clutter) along with the horizontal and vertical neighboring views from previous iterations.
% The warping follows a specific order: we always warp the pivot view first, followed by the vertical neighbor, and then the horizontal neighbor. 
% Change this because it's unknow why we first do vertical then horizontal
At each step, we only add pixels that were not already filled by earlier warps. Once the combined warped image is constructed, \textit{Masked-DiffReaging} is applied to inpainting the remaining unfilled regions. This strategy reduces the possibility of reconstructing overlapping regions independently and subsequently avoids blurring the details, as demonstrated in Fig.~\ref{fig:ablation} (``+Pivot+W+COR'').

\vspace{-0.05in}
\subsubsection{Optimization}
\label{subsub:optimization}

Here, we use the re-aged 2D images to supervise the optimization of the re-aged 3D representation $\mathcal{G}_r$. Specifically, we initialize $\mathcal{G}_r$ with the original representation $\mathcal{G}_o$, and optimize it by minimizing the following objective:

\vspace{-0.15in}
\begin{equation}
\label{eq:loss_final}
    \mathcal{L}_{\text{3D}} = \sum_{n \in \{1, \dots, N\} \setminus \{p\}} \bigg[\mathcal{L}_{\text{rec}} \big( R(\mathcal{G}_r, \vect{v}_n), \vect{L}_n \big) + \mathcal{L}_{\text{rec}} \big( R(\mathcal{G}_r, \vect{v}_p), \vect{L}_p \big) \bigg],
\end{equation}
\vspace{-0.15in}

\noindent where the first term is the reconstruction loss between the rendered and re-aged non-pivot view, and the second term is the same loss computed for the pivot view. Since we primarily use 3DGS as our representation, we adopt the same reconstruction loss used in 3DGS, which combines $L_1$ and SSIM.

We perform the optimization for 2,000 iterations and update the re-aged non-pivot images by performing the center-out re-aging every 400 steps. As shown in Fig.~\ref{fig:flow_warping_improvement}, as our 3D model improves, the flow quality improves which in turn improves the quality of the 3D re-aged representation.

% Where, $L_{c}=\left\| I_{\text{ren\_c}} - I_{\text{edi\_c}} \right\|^2$ and $L_{p}=\left\| I_{\text{ren\_p}} - I_{\text{edi\_p}} \right\|^2$.
% During the optimization process, the multi-view consistent reaging effects are effectively transferred to the 3D model, enabling realistic and coherent 3D reaging. 

% Recall our observation that flow estimated from the original renderings can be used to warp the re-aged image at the target age. This assumption generally holds because the re-aged 3D model typically maintains similar geometry to the original. However, when the target age differs significantly from the original, the flow derived from the original images becomes inaccurate for warping, as shown in iteration \#0 of Fig.~\ref{fig:flow_warping_improvement}. To address this issue, we iteratively update the flow. As the editing progresses, the age of the intermediate renderings gradually approaches the target age, resulting in increasingly accurate flow and, consequently, more precise reconstruction of the re-aged image, as demonstrated in Fig.~\ref{fig:flow_warping_improvement}.

% ----------------------------------------------------------------------------

\section{Results}
\label{sec:results}

We implement our approach in PyTorch and use the ADAM optimizer~\cite{Kingma15Adam} with default parameters for both the training of \textit{DiffReaging} and the optimization of the 3D model. \textit{DiffReaging} is fine-tuned on images with a resolution of $1024 \times 1024$ and a batch size of 4 for approximately three days using four NVIDIA A100 GPUs with 40GB of memory. For 3D model optimization, we use a single NVIDIA A5000 GPU with 24GB of memory. %All results presented in this section are based on 3DGS as the 3D representation, however, our approach is general and can handle any other optimizable 3D representation such as NeRF. 
We first present results for 2D re-aging, followed by the evaluation of our 3D face re-aging method.

% first present the re-aging results for 2D face images produced by our model against InstructPix2Pix~\cite{brooks2023instructpix2pix} and FRAN~\cite{zoss2022fran} in Sec.\ref{subseq:2d_face_reaging}. Subsequently, in Sec.\ref{subseq:3d_face_reaging}, we present 3D face editing results, including aging edits applied to 3D face models. A comparative analysis is then provided against five baseline methods, InstructGS2GS (IGS2GS)~\cite{igs2gs}, GaussianEditor (GE)~\cite{chen2024gaussianeditor}, DGE~\cite{chen2024dge}, and two hybrid methods, IGS2GS+DiffReaging (IGS2GS+) and GE+DiffReaging (GE+).

% ----------------------------------------------------------------------------
\begin{figure}[t]
% \vspace{-0.1in}
\centering
\includegraphics[width=0.9\linewidth, scale=1.0, angle=-0]{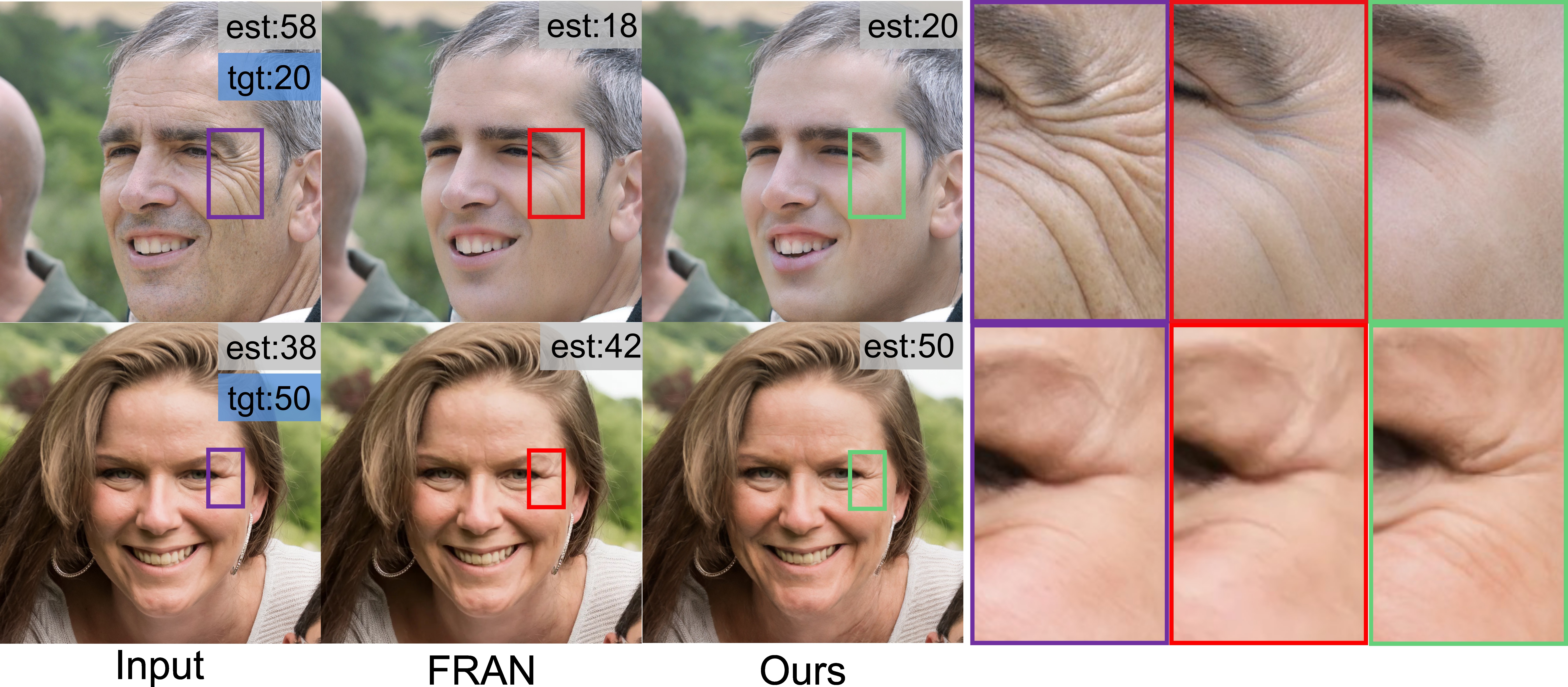}
\vspace{-0.1in}
\caption{Here, we show comparisons against FRAN. FRAN exhibits limitations in de-aging (top row), as it is not able to effectively remove deep wrinkles. Additionally, its editing capability diminishes when the target age is close to the source age, as shown in the bottom row. In contrast, our method achieves precise age control, enabling effective modification of the age attribute in input images.}
\vspace{-0.15in}
\label{fig:reaging_diffusion_2d2}
\end{figure}
% ----------------------------------------------------------------------------

\subsection{2D Face Re-Aging}
\label{subseq:2d_face_reaging}

\begin{table}[t] % not t or b or p
% \footnotesize
% \small
\caption{
\change{
Evaluating Age Accuracy. We measure age accuracy by calculating the difference between the estimated age of the re-aged results and the target ages. Lower values indicate better performance, with the best results highlighted in bold. \changenew{Additional results in 5-year increments provided in the supplementary document.}
}
}
\vspace{-0.1in}

% % ----------------------------------------------------------------------------
\centering
\resizebox{\columnwidth}{!}{%
\begin{tabular}{
p{0.05\textwidth}>
{\centering}p{0.015\textwidth}>
{\centering}p{0.015\textwidth}>
{\centering}p{0.015\textwidth}>
{\centering}p{0.015\textwidth}>
{\centering}p{0.015\textwidth}>
{\centering}p{0.015\textwidth}>
{\centering}p{0.015\textwidth}>
{\centering}p{0.015\textwidth}>
{\centering\arraybackslash}p{0.1\textwidth}
}
\hline
\hline
& \multicolumn{8}{c}{Age Error $\downarrow$} & \multirow{2}{*}{ID Score $\uparrow$}\\
\cmidrule(lr){2-9}
& 10 & 20 & 30 & 40 & 50 & 60 & 70 & 80 &\\
\hline

FADING  & 10.2 & 17.6 & 13.1 & 8.2 & 8.4 & 7.0 & 12.3 & 14.8 & 0.555$\pm$0.232 \\
IPix2Pix  & 9.4 & 12.3 & 11.3 & 12.4 & 9.3 & 13.1 & 17.8 & 13.6 & 0.604$\pm$0.248 \\
FRAN  & 4.2 & 5.1 & 3.9 & 4.9 & 6.3 & 5.1 & 11.2 & 13.3 & 0.675$\pm$0.199 \\
Ours  & \textbf{2.3} & \textbf{3.1} & \textbf{3.2} & \textbf{4.0} & \textbf{4.7} & \textbf{3.6} & \textbf{5.0} & \textbf{6.6}
 & \textbf{0.684$\pm$0.184} \\
\hline

\end{tabular}%
}

\label{tab:age_id_2d}
\vspace{-0.2in}
\end{table}

% ----------------------------------------------------------------------------
\begin{figure*}[htb]
\vspace{-0.1in}
\centering
\includegraphics[width=0.95\linewidth, scale=1.0, angle=-0]{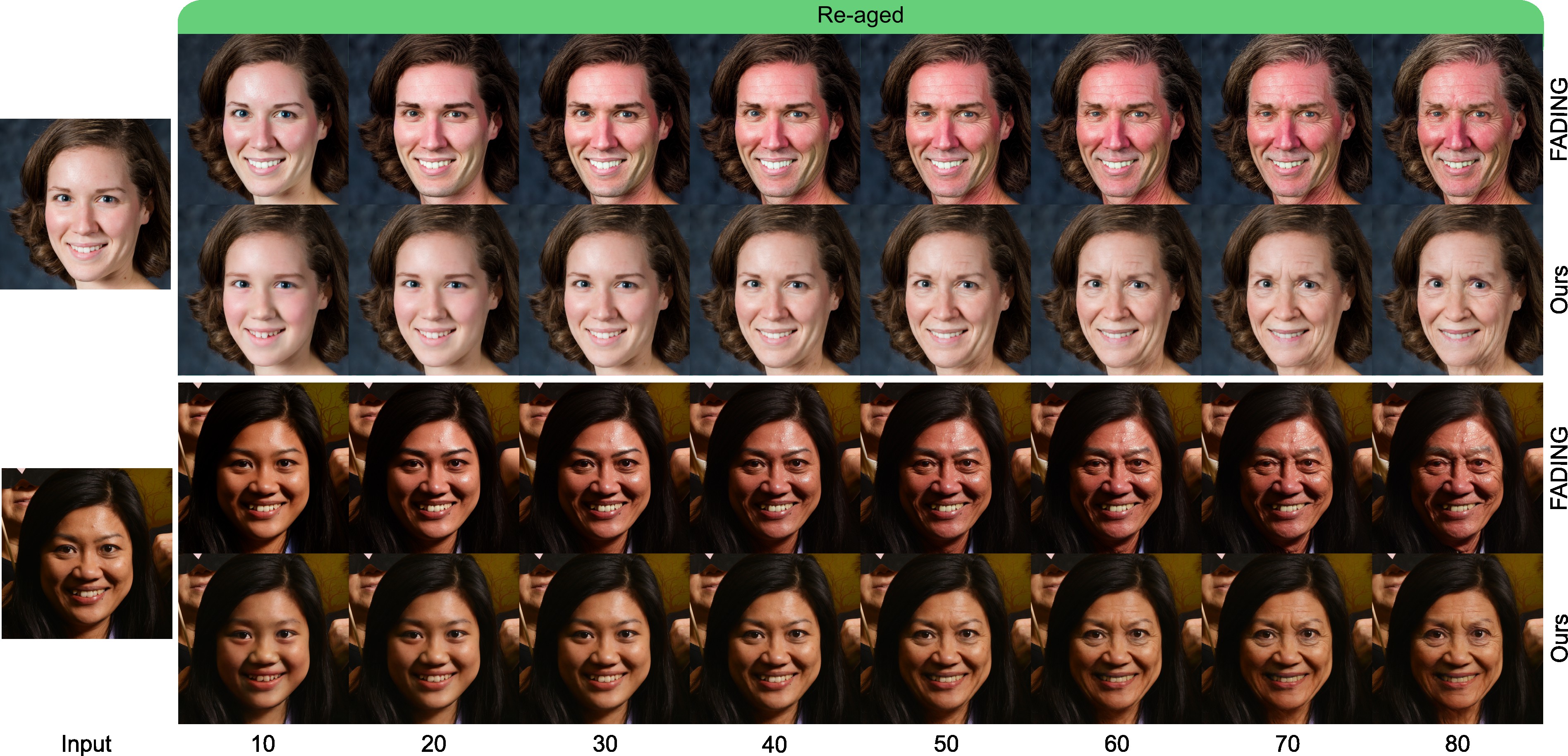}
\vspace{-0.1in}
\caption{Compared to FADING, our re-aging diffusion model consistently produces realistic age transformations while preserving the subject's identity. Furthermore, the age progression shows smooth transitions, exemplified by the gradual appearance and deepening of wrinkles as age increases.}
\vspace{-0.15in}
\label{fig:reaging_diffusion_2d_fading}
\end{figure*}
% ----------------------------------------------------------------------------

% ----------------------------------------------------------------------------
\begin{figure*}[htb]
% \vspace{-0.1in}
\centering
\includegraphics[width=0.9\linewidth, scale=1.0, angle=-0]{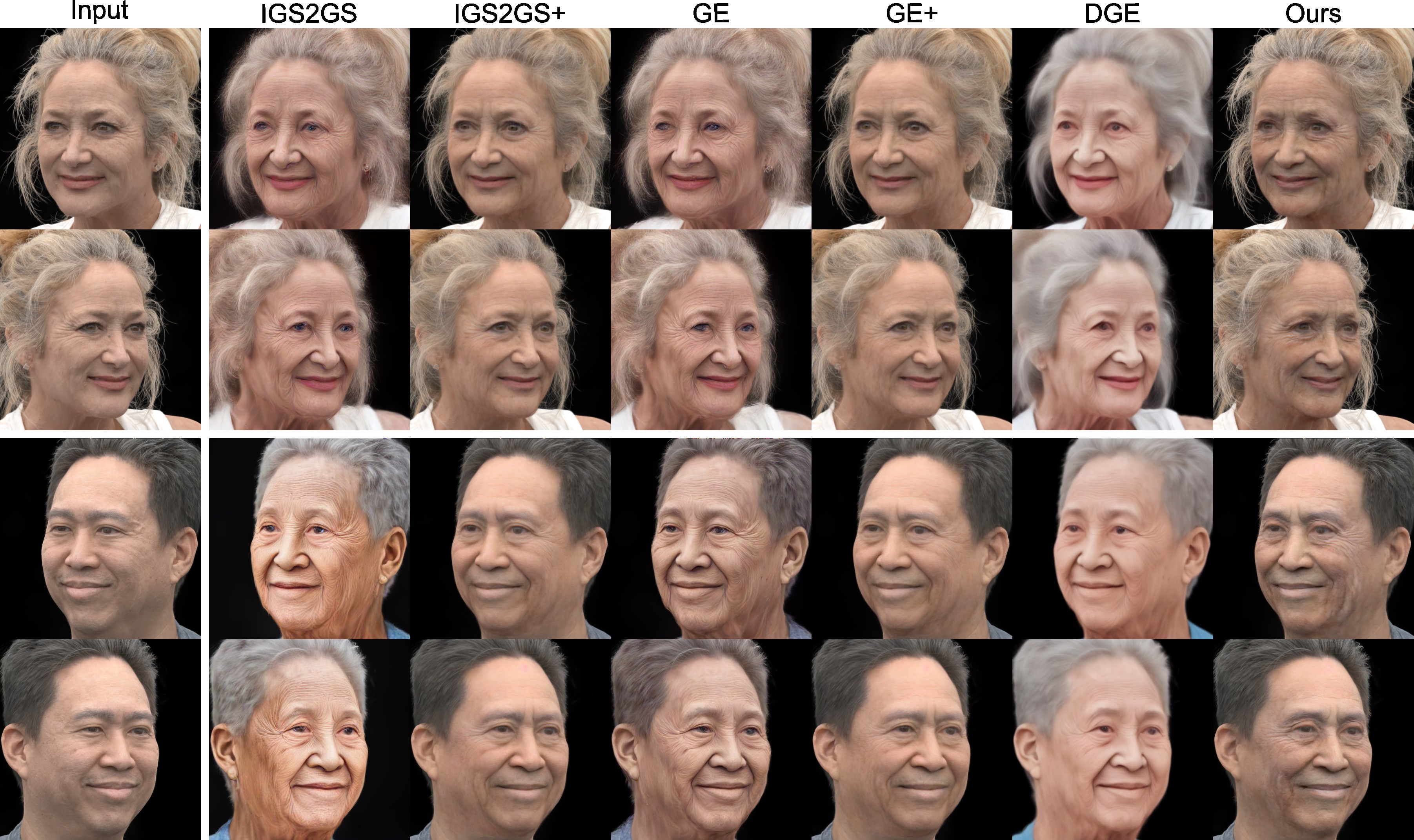}
\vspace{-0.1in}
\caption{We show comparisons against several methods. The baselines IGS2GS~\cite{igs2gs}, GE~\cite{chen2024gaussianeditor}, and DGE~\cite{chen2024dge}, all based on InstructPix2Pix~\cite{brooks2023instructpix2pix}, struggle to reage faces to 80 years, often introducing severe artifacts or over-smoothing due to reliance on feature-space constraints. Enhanced variants (IGS2GS+ and GE+) reduce artifacts using our re-aging diffusion model but still fail to reach the target age, often smoothing out critical features like wrinkles. For instance, they make the senior lady (about 60 years with deep wrinkles), appear younger than the input. In contrast, our method achieves accurate aging to 80 years, preserving fine age-related details and maintaining multi-view consistency.}
\vspace{-0.2in}
\label{fig:comparison}
\end{figure*}
% ----------------------------------------------------------------------------

% We compare the 2D re-aging results of our \textit{DiffReaging} with those of the general-purpose editing model InstructPix2Pix~\cite{brooks2023instructpix2pix} and diffusion-based face re-aging approach FADING~\cite{chen2023fading}.
We compare the 2D re-aging results of our \textit{DiffReaging} with those of the diffusion-based face re-aging approach FADING~\cite{chen2023fading}. As shown in Fig.~\ref{fig:reaging_diffusion_2d_fading}, FADING fails to accurately manipulate age-related attributes and struggles to preserve the subject's identity. In contrast, our method demonstrates precise control over the age attribute, successfully re-aging input images to target ages spanning from 10 to 80 years in 10-year increments. The results indicate that our re-aging diffusion model produces realistic outputs while maintaining identity across a wide range of target ages. 
% Additional visual comparisons with InstructPix2Pix~\cite{brooks2023instructpix2pix} are provided in the supplementary material.

We further compare our method against FRAN~\cite{zoss2022fran} in Fig.~\ref{fig:reaging_diffusion_2d2}. As shown, FRAN exhibits clear shortcomings, particularly when attempting to de-age the subject. For instance, in the top row, the 20-year-old output still retains deep wrinkles, indicating incomplete removal of age-related features. Additionally, FRAN struggles with age transitions when the target age is close to the source age. This is evident in the bottom row, where the model generates a 42-year-old appearance despite the target age being 50. In contrast, our method produces age-consistent results, demonstrating its effectiveness in modifying age while preserving identity.
\changenew{Moreover, our method, DiffReaging, is explicitly designed as the foundation for Masked-DiffReaging, enabling multiview-consistent and localized re-aging, which further expands the applicability of our method in high-fidelity, view-consistent face editing scenarios.}

In addition, we conduct a quantitative evaluation of age manipulation accuracy and identity preservation against FADING~\cite{chen2023fading},  InstructPix2Pix~\cite{brooks2023instructpix2pix} and FRAN~\cite{zoss2022fran}. Specifically, we randomly sample 1,000 unseen images from FFHQ~\cite{Karras2019stylegan2} and re-age each to target ages from 10 to 80 in 10-year increments. We assess the re-aged outputs using an age estimation model~\cite{rothe2018deep} and a face recognition model~\cite{deng2019arcface} to compute age prediction errors and identity similarity scores, respectively. The results are summarized in Table~\ref{tab:age_id_2d}. Our method consistently outperforms all baselines in terms of age consistency and identity similarity. 
%\changenew{Additional results are provided in the supplementary document.}

\vspace{-0.1in}

% ----------------------------------------------------------------------------
% ----------------------------------------------------------------------------

\subsection{3D Face Re-Aging}
\label{subseq:3d_face_reaging}

%on ten randomly selected subjects from the NeRSemble dataset~\cite{kirschstein2023nersemble}, as well as on

\begin{figure}[htb]
% \vspace{-0.05in}
\centering
\includegraphics[width=0.95\linewidth, scale=1.0, angle=-0]{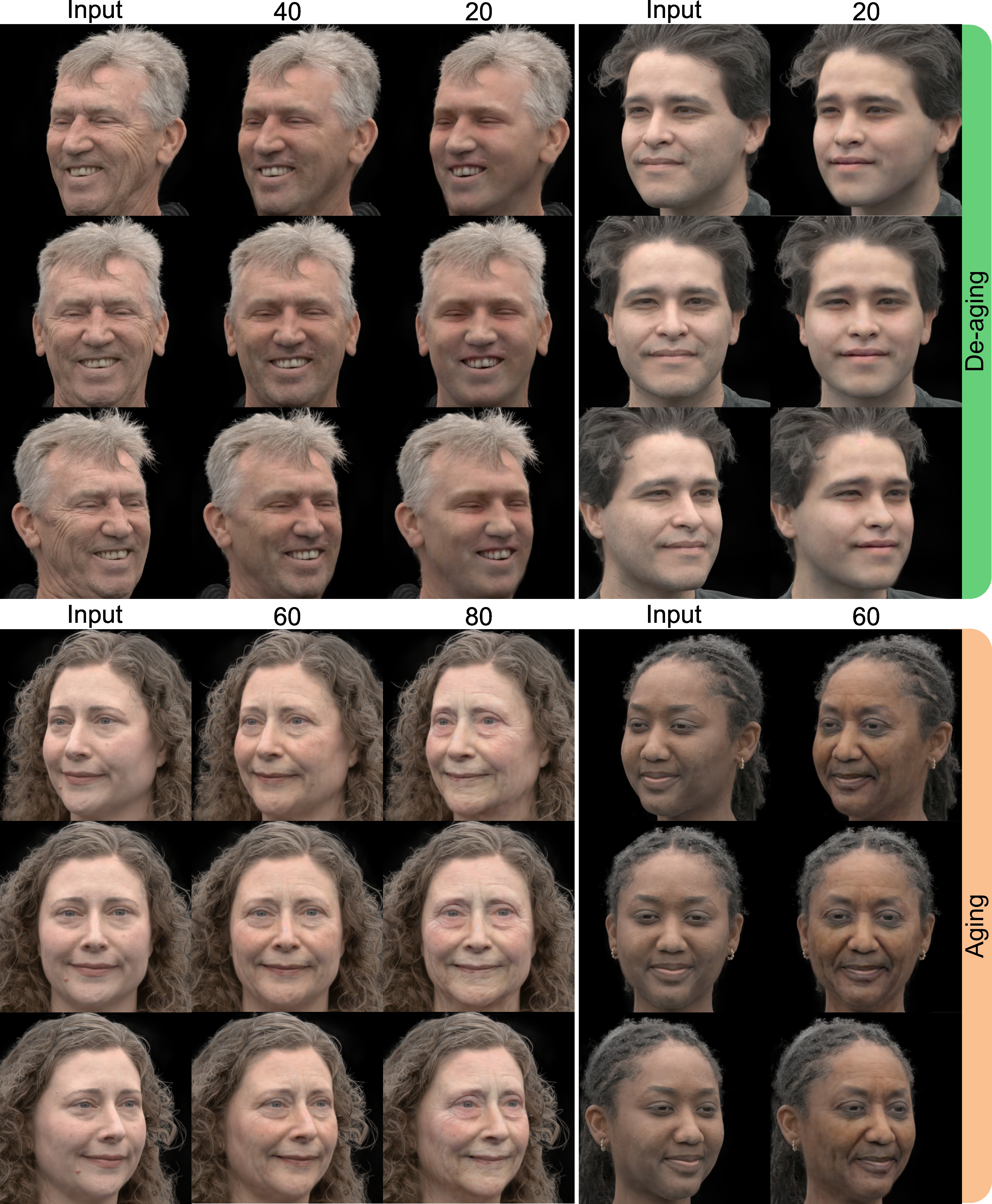}
\vspace{-0.1in}
\caption{We show de-aging results in the top row and aging results in the bottom row. As observed, age-related features gradually fade during de-aging and progressively emerge during aging. The re-aging process maintains strong multi-view consistency.}
% \vspace{-0.1in}
\label{fig:reging_3d}
\end{figure}

We evaluate 3D face re-aging results \changenew{ten subjects from our internal dataset. Each subject in our internal dataset was captured using a light stage setup~\cite{debevec2000acquiring} with 76 cameras.} These images are used to optimize a 3DGS model \egsrcameraready{using the standard process with densification and pruning}, which serves as the input representation $\mathcal{G}_o$ for our method. As shown in Fig.~\ref{fig:teaser}, our approach enables fine-grained age manipulation with strong identity preservation and multi-view consistency. It supports smooth transitions from ages 20 to 80, capturing gradual changes such as wrinkle formation. Additional results in Fig.~\ref{fig:reging_3d} show both aging and de-aging across a diverse set of subjects, demonstrating the robustness of our method.

\vspace{-0.1in}
\paragraph{Qualitative Comparisons} We compare against five baselines: InstructGS2GS (IGS2GS)~\cite{igs2gs}, GaussianEditor (GE)~\cite{chen2024gaussianeditor}, DGE~\cite{chen2024dge}, and two hybrid methods, IGS2GS+\textit{DiffReaging} (IGS2GS+) and GE+\textit{DiffReaging} (GE+), in which we integrate our \textit{DiffReaging} model into the original algorithms. Notably, neither IGS2GS/IGS2GS+ nor our approach performs densification or pruning during the editing process. In contrast, both GE/GE+ and DGE apply explicit densification, following the configurations recommended in their official implementations: every 500 iterations for GE/GE+ and every 100 iterations for DGE.

% ----------------------------------------------------------------------------
\begin{figure}[htb]
% \vspace{-0.1in}
\centering
\includegraphics[width=1.0\linewidth, scale=1.0, angle=-0]{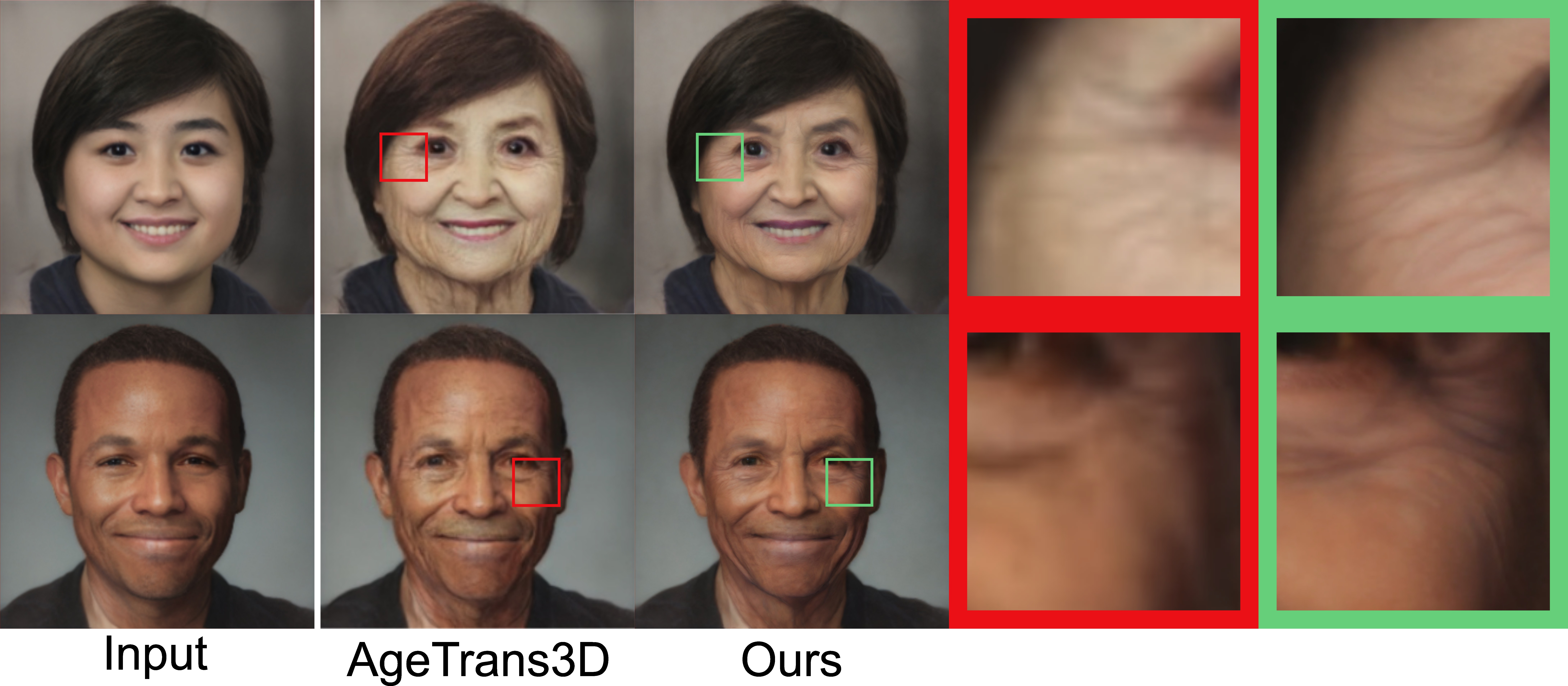}
\vspace{-0.2in}
\caption{\egsrcameraready{We present a comparison with 2D re-aging of AgeTrans3D~\cite{li2024age}. Our method generates higher-resolution results and more realistic age-related details.}}
\vspace{-0.15in}
\label{fig_sup:reaging_diffusion_2d_lglg}
\end{figure}
% ----------------------------------------------------------------------------

% ----------------------------------------------------------------------------
% ----------------------------------------------------------------------------

As shown in Fig.~\ref{fig:comparison}, IGS2GS, GE, and DGE, each based on InstructPix2Pix, struggle to re-age subjects to 80 years, often introducing severe artifacts. In particular, DGE produces over-smoothed results, likely due to its reliance on feature-space consistency, which does not always translate to consistency in image space. While the enhanced variants IGS2GS+ and GE+ incorporate our diffusion-based re-aging model and thus exhibit fewer artifacts than their original versions, they tend to overblur age-related details and fail to achieve the desired target age. Notably, in the case of the senior woman shown at the top (60 years old with deep wrinkles), these methods smooth out key age cues in the re-aged image, making the output appear even younger than the input. In contrast, our method accurately reaches the target age of 80, preserving age-related features such as wrinkles. \changenew{We emphasize that the final renderings produced by all methods are inherently view-consistent, as they are generated from optimized 3D representations. The key difference is that existing methods produce re-aged 2D target images that are not consistent across views. Optimizing a 3D model under such conflicting supervision inevitably averages out high-frequency age cues, resulting in over-smoothed and blurry renderings.}

\egsrcameraready{We also provide comparisons with AgeTrans3D~\cite{li2024age}. AgeTrans3D performs single-image 2D re-aging using a diffusion model while additionally considering biophysical skin properties~\cite{iglesias2015biophysically}, and lifts the re-aged image to 3D. However, its 2D re-aging model is trained from scratch as a low-resolution pixel-space diffusion model, whereas our DiffReaging fine-tunes a pretrained image-editing latent diffusion model with strong priors for photorealistic image generation. This enables our method to produce higher-resolution re-aging results with finer facial details. Since the AgeTrans3D code is not publicly available, we compare against two representative examples from their paper that showcase their best reported results. As shown in Fig.~\ref{fig_sup:reaging_diffusion_2d_lglg}, our method produces higher-resolution results with finer age-related details, including natural wrinkles around the eyes. However, unlike AgeTrans3D, we do not model the biophysical skin appearance changes associated with aging, such as the natural lightening of skin tone.}

% ----------------------------------------------------------------------------
\begin{figure}[htb]
% \vspace{-0.05in}
\centering
\includegraphics[width=0.95\linewidth, scale=1.0, angle=-0]{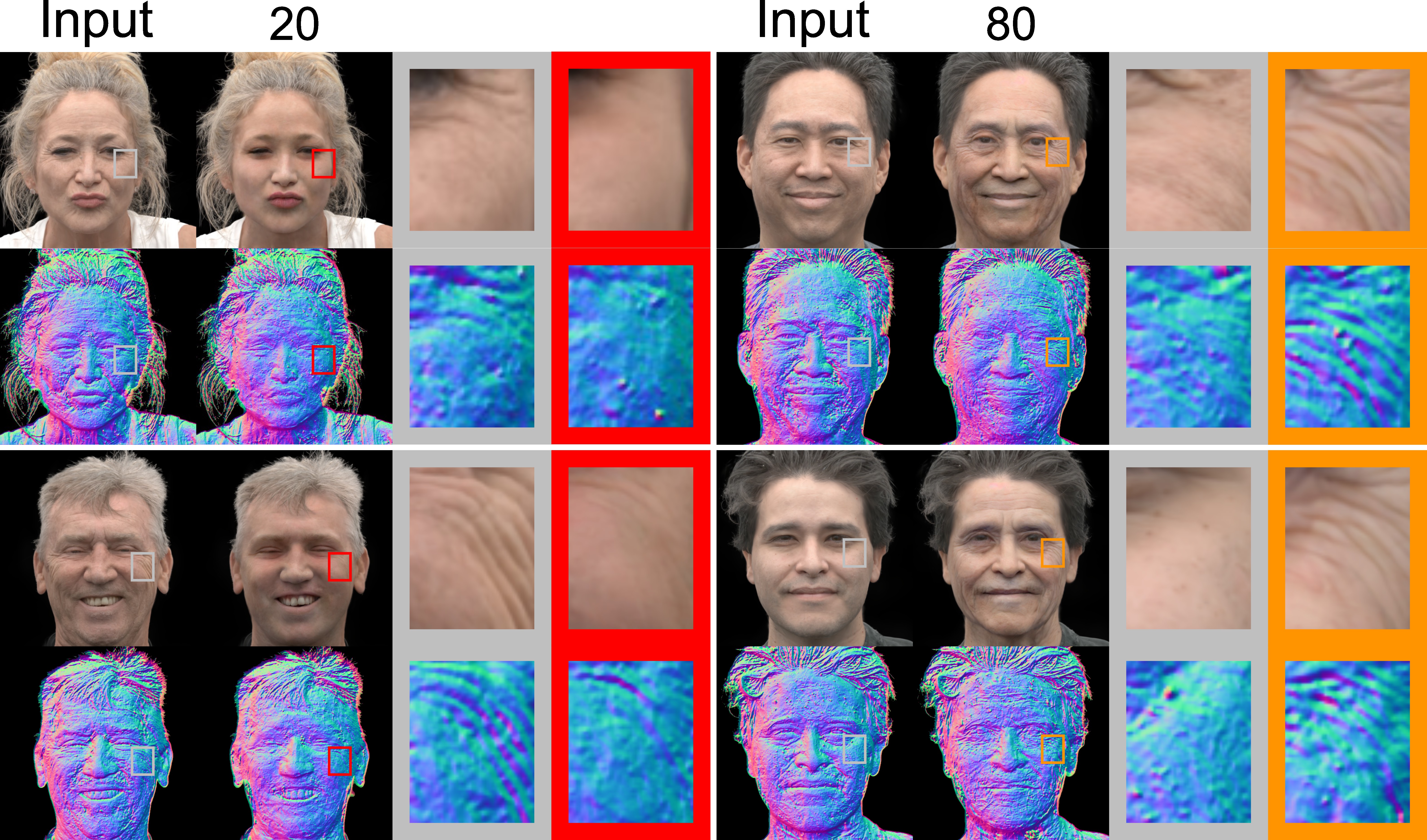}
\vspace{-0.1in}
\caption{Geometry editing effect. Our method enables geometric changes, as demonstrated through normal maps extracted from 3DGS~\cite{3dgaussian2023} models.}
\vspace{-0.1in}
\label{fig:geometry}
\end{figure}
% ----------------------------------------------------------------------------

\changenew{
To further demonstrate the geometric editing capability of our method, we extract normal maps from 3DGS~\cite{3dgaussian2023} models, as illustrated in Fig.~\ref{fig:geometry}. The visualizations show that as the subjects de-aged or aged, age-related geometric details—such as wrinkles, and skin sagging—emerge clearly in the normal maps, indicating that our method introduces meaningful geometry changes consistent with natural aging.
% The visualizations reveal that, as the subject is re-aged across different target ages, age-related geometric features such as wrinkles, fine creases, and skin sagging become clearly manifested in the corresponding normal maps. This indicates that our approach not only modifies texture appearance but also introduces meaningful geometric variations consistent with the natural aging process.
}

% ----------------------------------------------------------------------------
\begin{table}[t] % not t or b or p
% \footnotesize
% \small
\caption{
\change{
Age Accuracy Comparison. We measure age errors relative to target ages. Our method achieves low errors across all targets, while the enhanced baselines show significant underestimation for ages 70 and 80, often regressing to about 60, confirming their lack of age consistency.
}
}
\vspace{-0.1in}

% % ----------------------------------------------------------------------------
\centering
\resizebox{\columnwidth}{!}{%
\begin{tabular}{p{0.06\textwidth}>
{\centering}p{0.02\textwidth}>
{\centering}p{0.02\textwidth}>
{\centering}p{0.02\textwidth}>
{\centering}p{0.02\textwidth}>
{\centering}p{0.02\textwidth}>
{\centering}p{0.02\textwidth}>
{\centering}p{0.02\textwidth}>
{\centering}p{0.02\textwidth}>
{\centering\arraybackslash}p{0.06\textwidth}}
\hline
\hline
& \multicolumn{9}{c}{Age Error $\downarrow$}\\
\cmidrule(lr){2-10}
& 10 & 20 & 30 & 40 & 50 & 60 & 70 & 80 & Avg. \%\\
\hline
IGS2GS+  & 4.8 & 5.4 & 4.6 & 5.4 & 6.9 & 4.8 & 11.0 & 19.5 & 20.7\% \\
GE+  & 4.4 & 4.8 & 6.3 & 6.0 & 7.7 & 5.1 & 12.1 & 19.4 & 21.2\% \\
Ours  & \textbf{3.0} & \textbf{3.8} & \textbf{3.1} & \textbf{4.6} & \textbf{5.6} & \textbf{2.9} & \textbf{8.4} & \textbf{9.6} & \textbf{13.9\%} \\
\hline

\end{tabular}%
}

\label{tab:age_3d}
% \vspace{-0.1in}
\end{table}

% ----------------------------------------------------------------------------
\begin{table}[htb] % not t or b or p
% \footnotesize
% \small
\caption{
Identity Preservation (ID Score) Comparison. Our method yields higher average ID scores and lower variance, reflecting stronger identity preservation.
% \mingming{Rotate the table and put all methods in the first row with their scores in the second row to save some space.
}
\vspace{-0.1in}

% % ----------------------------------------------------------------------------
\centering
\resizebox{\columnwidth}{!}{%
\begin{tabular}{p{0.075\textwidth}>
{\centering}p{0.075\textwidth}>
{\centering}p{0.075\textwidth}>
{\centering}p{0.075\textwidth}>
{\centering\arraybackslash}p{0.085\textwidth}}
\hline
\hline
& IGS2GS & GE & DGE & Ours \\
\hline

ID Score $\uparrow$  & 0.218$\pm$0.042 & 0.149$\pm$0.047 & 0.206$\pm$0.026 & \textbf{0.671$\pm$0.019}\\

\hline

\end{tabular}%
}

\label{tab:id_3d}
\vspace{-0.1in}
\end{table}

% ----------------------------------------------------------------------------

\emph{Quantitative Comparisons.} To numerically assess the age control capability of our approach, we select ten subjects, each exhibiting four distinct facial expressions: neutral, pucker face, closed-mouth smile (smile without teeth), and open-mouth smile (smile with teeth). Each subject is re-aged to the target ages of 10, 20, 30, 40, 50, 60, 70, and 80 years. From the final re-aged models, we render 30 views per subject, covering yaw angles from -60\textdegree{} to 60\textdegree{} with a fixed pitch angle of 0\textdegree{}. 
We compare against IGS2GS, GE, and DGE using ArcFace~\cite{deng2019arcface}, a face recognition method, to compute identity similarity scores (ID Scores) between original and re-aged images. As shown in Table~\ref{tab:id_3d}, our method achieves higher average ID scores with lower variance, indicating better identity preservation. For enhanced baselines (IGS2GS+ and GE+), we report age errors relative to target ages in Table~\ref{tab:age_3d}. As seen, the enhanced baselines show larger errors than ours, particularly for age 80, demonstrating their inability to produce age-consistent results.

\changenew{Additional comparisons on ten randomly selected subjects from NeRSemble dataset~\cite{kirschstein2023nersemble}, as well as further results with 3D-aware diffusion models such as LRM~\cite{hong2023lrm,xu2024instantmesh} and One-2-3-45~\cite{liu2023one,liu2024one}, are provided in the supplementary document.}
% ----------------------------------------------------------------------------

% ----------------------------------------------------------------------------
\begin{figure}[t]
% \vspace{-0.1in}
\centering
\includegraphics[width=1.0\linewidth, scale=1.0, angle=-0]{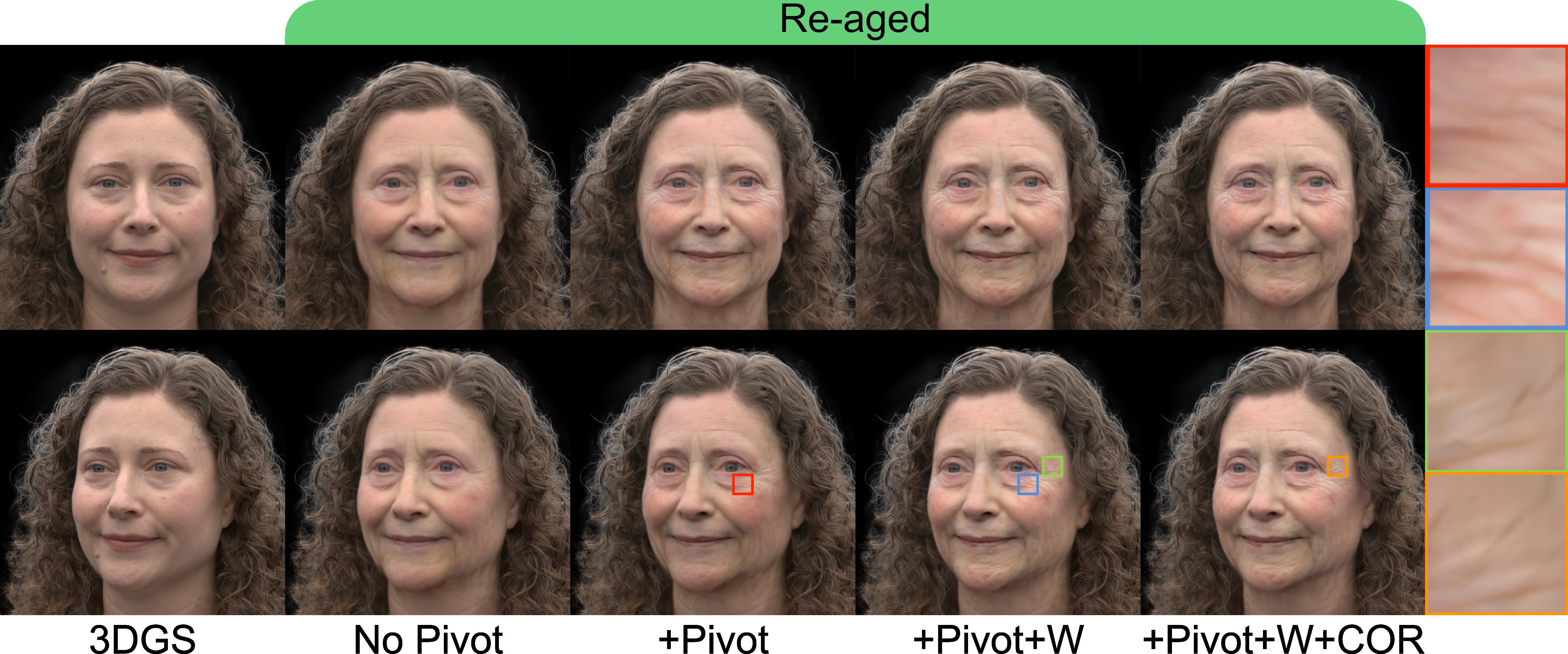}
\vspace{-0.2in}
\caption{Ablation Study. The "+Pivot" variation ensures that the pivot view of the 3D model is effectively edited to match the target attributes. The "+Pivot+W" variation facilitates the propagation of information from the pivot view to other views, as evidenced by the blue box, which demonstrates superior results compared to the red box region. Finally, the "+Pivot+W+COR" variation further reduces the randomness in the inpainted pixels, as highlighted by the orange box, which exhibits sharper and more refined details of the ear, compared to the green box region.}
% \vspace{-0.1in}
\label{fig:ablation}
\end{figure}
% ----------------------------------------------------------------------------

% ----------------------------------------------------------------------------
\begin{figure}[htb]
% \vspace{-0.1in}
\centering
\includegraphics[width=1.0\linewidth, scale=1.0, angle=-0]{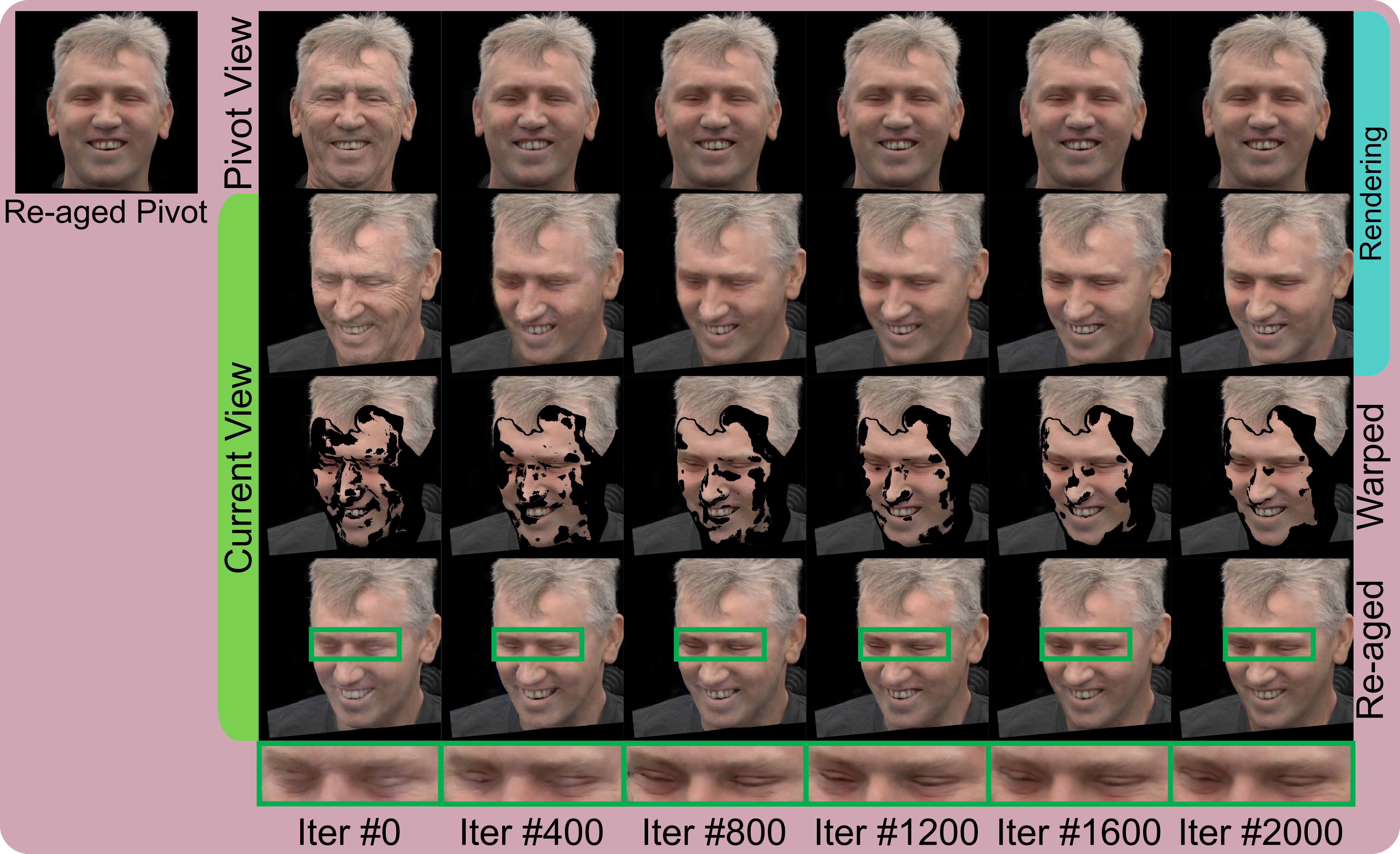}
\vspace{-0.2in}
\caption{We present intermediate results from the iterative re-aging process. As the iterations progress, the accuracy of flow warping improves because the rendered images used to compute the flow increasingly resemble the target age. This leads to more precise reconstruction of the re-aged image, as highlighted by the green boxes.}
\vspace{-0.2in}
\label{fig:flow_warping_improvement}
\end{figure}
% ----------------------------------------------------------------------------

\emph{Ablation Study.} To assess the contribution of each component in our proposed method, we conduct an ablation study by progressively incorporating key modules. We begin with the ``No Pivot'' variant, which is equivalent to the IGS2GS+ baseline. We then introduce the pivot editing reference along with the pivot consistency loss (full loss in Eq.~\ref{eq:loss_final}), resulting in the ``+Pivot'' variant. Next, we incorporate the warping strategy to propagate the pivot edit across novel views, forming the ``+Pivot+W'' variant. Finally, we introduce the Center-Out Re-aging (COR) strategy to obtain the full model, denoted as ``+Pivot+W+COR''.
Comparative results are presented in Fig.~\ref{fig:ablation}. The ``+Pivot'' model ensures accurate editing of the pivot view, aligning it closely with the target aging attributes. The ``+Pivot+W'' variant enables effective propagation of this edit to other viewpoints, as illustrated by the blue inset, which shows improved details compared to the red inset. Lastly, the full ``+Pivot+W+COR'' model further enhances detail preservation by reducing the independent reconstruction of overlapping regions in different views. This is particularly evident in the orange inset, which exhibits sharper and more refined details compared to the corresponding green inset. \changenew{The corresponding numerical ablation study is provided in the supplementary document.}
% \changenew{Furthermore, we would like to highlight that IGS2GS+ in Fig.~\ref{fig:comparison} and Table~\ref{tab:age_3d} effectively serves as an ablation baseline without the full propagation module, which demonstrates the effectiveness of the propagation strategy.}

We further evaluate the necessity of iterative re-aging rather than a single-step transformation. As shown in Fig.~\ref{fig:flow_warping_improvement}, flow warping becomes more accurate with each iteration, as the rendered images used to compute flow progressively align with the target age. This leads to a more precise reconstruction of re-aged features, as highlighted in the green boxes.

% % ----------------------------------------------------------------------------

% \emph{3D Generative Approaches.} Given multi-view images, alternative 3D generative methods such as LRM~\cite{hong2023lrm,xu2024instantmesh} and general-purpose 3D generators like One-2-3-45~\cite{liu2023one,liu2024one}, can potentially be adopted for 3D face re-aging. Specifically, this can be achieved by first applying 2D re-aging using our DiffReaging model, followed by employing these 3D generators to reconstruct a 3D face from the re-aged 2D images.
% As illustrated in Fig.~\ref{fig:generative3d}, the left side shows six re-aged views produced by DiffReaging, which serve as inputs to LRM~\cite{hong2023lrm,xu2024instantmesh} and One-2-3-45~\cite{liu2023one,liu2024one}. The resulting 3D outputs, displayed on the right, demonstrate that these methods struggle to generate high-quality 3D facial representations. A likely reason for this limitation is that these models are not specifically optimized for 3D face generation.

% \input{5_limitations}
\section{Conclusion, Limitations, and Future Work}

We have presented a novel framework for realistic 3D face re-aging that produces identity-preserving results with fine-grained age-related details. At the core of our approach is a diffusion-based 2D re-aging model, \textit{DiffReaging}, trained on synthetically generated re-aged image pairs. To produce multi-view consistent re-aged images, we introduce a propagation strategy that re-ages a single pivot view and reconstructs the remaining views through warping and our proposed \textit{Masked-DiffReaging} mechanism. \textit{Masked-DiffReaging} leverages the iterative nature of \textit{DiffReaging} to ensure coherence between the reconstructed regions and existing warped content. Moreover, we propose a center-out re-aging method to avoid redundant reconstruction of overlapping regions by progressively expanding from the pivot in concentric layers. The resulting re-aged images supervise the optimization of the re-aged 3D representation, enabling age-consistent and identity-preserving 3D re-aged results. Extensive experiments demonstrate that our method outperforms existing 3D editing approaches both visually and numerically.

\begin{figure}[t]
% \vspace{-0.1in}
\centering
\includegraphics[width=1.0\linewidth, scale=1.0, angle=-0]{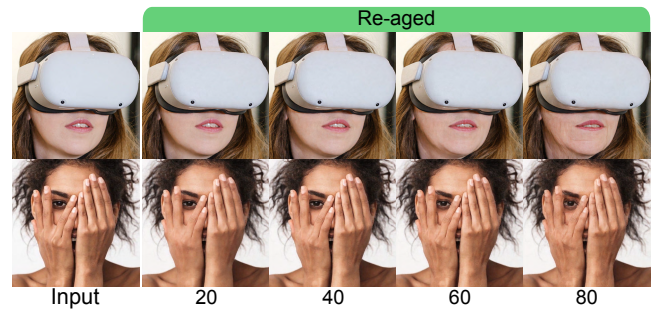}
\vspace{-0.2in}
\caption{Here, we highlight the limitations of DiffReaging on portrait images with occlusions. While age-related features gradually emerge during the process of re-aging, such as wrinkles forming on the lower face of the fair-skinned subject and around the eyes of the darker-skinned subject, visible artifacts also appear.}
\vspace{-0.2in}
\label{fig:occlusion}
\end{figure}
% ----------------------------------------------------------------------------

While \textit{DiffReaging} achieves state-of-the-art results, it has limitations when re-aging faces with occlusions, as shown in Fig.~\ref{fig:occlusion}, which may also affect downstream 3D face re-aging. Additionally, it currently does not account for aging in other body parts like hair or hands. 
\egsrcameraready{Our method focuses on the facial region, which is often the most challenging and identity-critical component of re-aging in film production. We therefore restrict editing to masked facial regions while leaving non-facial regions unchanged, naturally preserving aging-irrelevant accessories such as earrings or necklaces. Severe facial occlusions, such as large sunglasses, remain challenging cases for future work. Moreover, our method does not explicitly model age-related biophysical albedo changes~\cite{iglesias2015biophysically}, such as skin lightening, reduced redness, or changes in translucency. While the approach of AgeTrans3D~\cite{li2024age} incorporates these effects, it trains a pixel-space diffusion model from scratch and consequently produces results with fewer age-related details (see Fig.~\ref{fig_sup:reaging_diffusion_2d_lglg}). We believe incorporating a biophysical appearance model into our system is a promising direction and leave it for future investigation.}

\egsrcameraready{Finally, our method targets high-fidelity re-aging of captured multi-view 3D faces, where multiple views provide reliable identity and geometry cues. Removing this requirement is an interesting direction for future work. For example, one could combine our re-aging propagation framework with single-image or sparse-view 3D-aware diffusion methods, such as Xiang et al.~\cite{xiang20233d}, to hallucinate missing viewpoints before optimization.}

% where we aim to explore effective methods for incorporating re-aging effects into hair while maintaining consistency with the overall re-aging process.
\section{Acknowledgements}

We sincerely thank the anonymous reviewers for their valuable feedback and constructive suggestions. Additionally, portions of this research were conducted with the advanced computing resources provided by Texas A\&M High Performance Research Computing.

\vfill\eject
\clearpage

\raggedbottom
\bibliographystyle{eg-alpha-doi}
\bibliography{egbib}

\pagebreak

\end{document}